\documentclass[11pt,draftcls,onecolumn]{IEEEtran}

\usepackage[pdftex]{graphicx}
\graphicspath{{./fig/}}
\DeclareGraphicsExtensions{.pdf,.jpeg,.png}
\usepackage{amsmath,epsfig}
\usepackage{mathtools}
\usepackage{amssymb}
\usepackage{subfig}
\usepackage{multirow}
\usepackage{bm}
\usepackage{booktabs}
\usepackage[colorlinks=true,linkcolor=blue,citecolor=blue]{hyperref}
\usepackage[dvipsnames]{xcolor}

 \newcommand{\xd}[1]{{\leavevmode\color{black}{#1}}}
 \newcommand{\htwai}[1]{{\color{black}#1}}
 
 \newcommand{\dt}[1]{{\leavevmode\color{black}{#1}}}

\usepackage{amsthm}
\newtheorem{example}{Example}
\newtheorem{problem}{Problem}
\newtheorem{taggedproblemx}{Problem}
\newenvironment{taggedproblem}[1]
 {\renewcommand\thetaggedproblemx{#1}\taggedproblemx}
 {\endtaggedproblemx}

\theoremstyle{definition}
\newtheorem{approach}{Approach}

\begin{document}
\title{From topology learning to graph generation:\\ A unifying perspective}
\author{Xiaowen Dong*, Hoi-To Wai*, Siheng Chen, Laura Toni, and Dorina Thanou*
\thanks{Xiaowen Dong is with University of Oxford (e-mail: \href{mailto:xdong@robots.ox.ac.uk}{xdong@robots.ox.ac.uk}); Hoi-To Wai is with The Chinese University of Hong Kong (e-mail: \href{mailto:htwai@cuhk.edu.hk}{htwai@cuhk.edu.hk}); Siheng Chen is with Shanghai Jiao Tong University (e-mail: \href{mailto:sihengc@sjtu.edu.cn}{sihengc@sjtu.edu.cn}); Laura Toni is with University College London (e-mail: \href{mailto:l.toni@ucl.ac.uk}{l.toni@ucl.ac.uk}); Dorina Thanou is with EPFL (e-mail: \href{mailto:dorina.thanou@epfl.ch}{dorina.thanou@epfl.ch}). *Authors contributed equally.}}
\maketitle

\begin{abstract}
Learning graph structures from data is a fundamental problem that spans a wide range of signal processing and machine learning tasks. While significant effort has been made to tackle the problem, \dt{existing research has largely evolved along two parallel directions. The first seeks to infer the topology of an individual graph from observations supported on it, whereas the second seeks to learn a generative distribution from observed graph instances, enabling the sampling of new graphs.} 
\dt{This review presents a unified framework that connects these formulations by viewing them as inverse problems of a common generation process for graph data. We review the major methodologies within this framework, highlight their relationships, strengths, and limitations, and identify opportunities for integrating ideas across paradigms. By bridging graph topology learning and graph generation, this review provides a broader cross-disciplinary  perspective on the field and outlines promising directions for future research.}

\end{abstract}

\section{Introduction}
\label{sec:intro}

Learning the underlying structure from data plays a fundamental role in subsequent processing and analysis tasks. In many domains, this structure is naturally represented as a graph, which provides a flexible mathematical framework for modeling pairwise or higher-order relationships among entities. While graphs are often treated as given, in many practical settings they are only partially observed or entirely unknown, making structure learning a fundamental challenge. Broadly, graph structure learning can be viewed from two complementary perspectives. The first seeks to learn the topology of an individual graph by estimating its nodes and edges from data. The second aims to learn a graph generative model that captures the distribution over a collection of graphs, enabling the characterization and synthesis of graph ensembles, \dt{as well as sampling of new graph instances. Throughout this article, we use the term ``graph structure learning'' to refer to learning either  the topology of individual graphs directly or the distributions governing collections of graphs indirectly. }

\begin{figure}[t]
    \centering
    \includegraphics[width=1.0\linewidth]{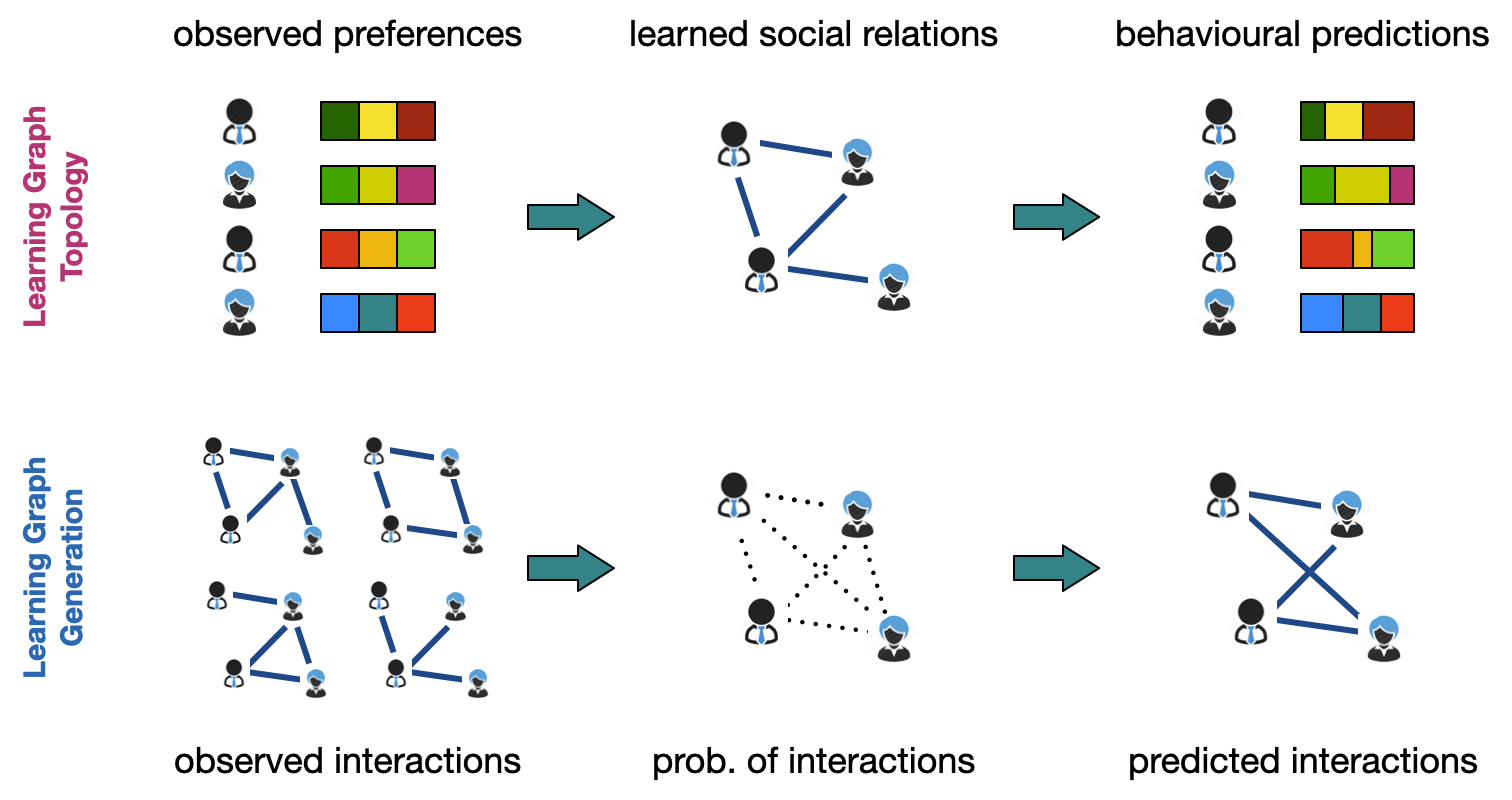}
    \caption{An illustration of the two problem instances of graph structure learning in social networks.}
    \label{fig:social}
\end{figure}

\dt{The two perspectives  differ not only in the final objective, but also in the nature of the observed data. In the first formulation, which we term ``learning graph topology'', observations typically consist of node observations supported on an unknown graph, whereas in the second, which we term ``learning graph generation'', the observations are graph samples drawn from an underlying graph distribution. This distinction is illustrated through the following social network example} 
(Fig. \ref{fig:social}). In the top row, we observe preferences or behaviors of a group of individuals \dt{in the form of graph signals}, and aim to learn the graph topology that characterizes the social relations among them. The social network topology can be used to identify key individuals in the network, or facilitate downstream tasks such as user-level predictions. In the bottom row, we observe social interactions among groups of individuals \dt{in the form of graph samples}, and aim to learn a probability distribution that governs the generation of such interactions. The learned distribution can then be used to predict future interactions or generate an entirely new social network topology. 
 
\begin{figure}[t]
    \centering
    \includegraphics[width=0.8\linewidth]{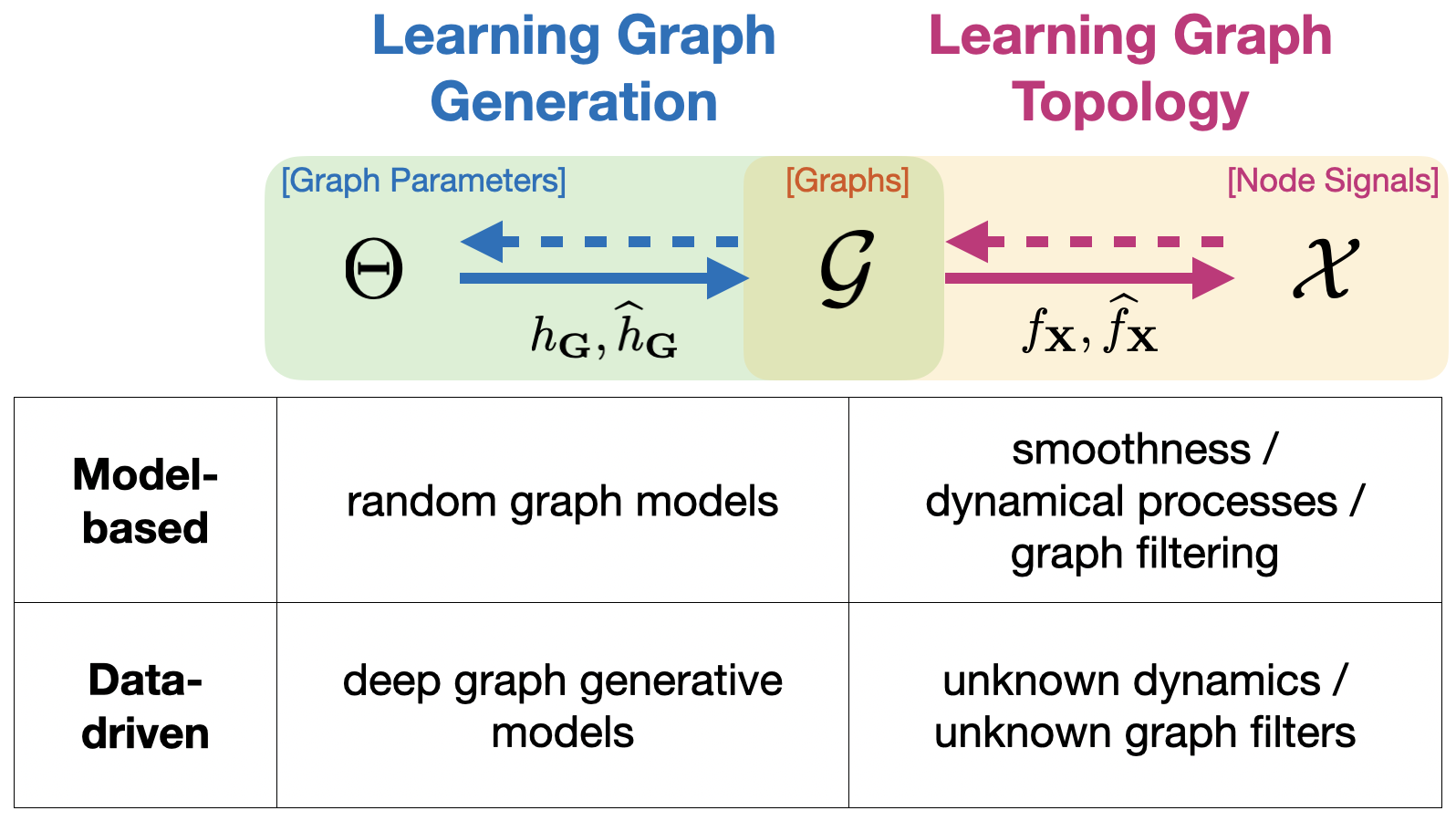}
    \caption{An illustration of the two families of approaches to graph structure learning.}
    \label{fig:paradigm}
\end{figure}

To provide a unified perspective, this review introduces a two-stage generation process of graph-related data in Fig.~\ref{fig:paradigm} and defines the learning problems as the inverse problems of each stage. We move from  left to  right  in the figure, where we use $\Theta$ to denote the parameter space for the graph generation process, ${\cal G}$ to denote the space of graph topologies (represented by an adjacency matrix $\mathbf{W} \in \mathbb{R}^{N \times N}$ for example), and ${\cal X}$ to denote the space of node observations, i.e., graph signals (represented by a vector $\mathbf{x} \in \mathbb{R}^N$). The first stage of the generation process is defined by a mapping $h: \Theta \to {\cal G}$ from the parameter space to the graph topology space, which describes how a graph is generated from its underlying parameters. For $\theta \in \Theta$, $h(\theta)$ typically refers to a probability distribution supported on ${\cal G}$. The second stage is defined by a mapping $f: {\cal G} \to {\cal X}$ from the graph topology space to the graph signal space, which describes how node observations are generated from a given graph. \dt{Together, these two mappings describe a common generation process for graph-structured data.} 
Within this framework, the two main graph structure learning problems naturally arise as inverse problems of the corresponding generation stages. In particular, \emph{learning graph topology} corresponds to inferring $G$ from observed $\mathbf{x}$ via the inverse of $f$, while \emph{learning graph generation} corresponds to inferring $\theta$ from observed $G$ via the inverse of $h$.

Existing approaches for graph structure learning can be broadly categorized into two  families: 1) model-based learning and 2) data-driven learning\footnote{In this review, by model-based learning, we mean that there exists a known model (e.g., statistical or physical) behind the data-generating process, and the task is to learn the parameters associated with this model; by data-driven learning, we assume that no such model exists as prior knowledge, and the task is to approximate the data-generating process and make predictions using function approximators such as neural networks.}. For graph topology learning, the two approaches differ by whether the $f$-map is assumed to be known beforehand (denoted by $f_\mathbf{X}$ in Fig.~\ref{fig:paradigm}) or not (denoted by $\widehat{f}_\mathbf{X}$). In the model-based setting, $f_\mathbf{X}$ is typically derived from statistical or physical principles. 
This leads to a rigorous mathematical formulation that often enables strong identifiability guarantees, but its applicability is inherently limited by the validity of the underlying modeling assumptions.
In contrast, data-driven approaches make minimal assumptions about the data-generating process and instead learn an implicit $\widehat{f}_{\mathbf{X}}$ directly from data.
They are more adaptive as they naturally allow learning to be guided by a downstream task. For example, the growing literature on \textit{graph structure learning} and \textit{graph rewiring}, as termed by the graph machine learning (GML) community, aims to refine an existing graph topology to better serve downstream tasks such as node or graph classification. A similar trade-off exists for learning graph generation. Herein, model-based and data-driven learning are then distinguished by whether the $h$-map is assumed to be known beforehand (denoted by $h_\mathbf{G}$ in Fig.~\ref{fig:paradigm}) or not (denoted by $\widehat{h}_\mathbf{G}$). The former typically relies on chosen random graph models, while the latter facilitates learning via latent node representations, which are then used for graph generation. As a consequence, it often enables the learning of a probability distribution, instead of merely a point estimate. 
Similar to graph topology learning, this increased flexibility comes at the cost of reduced interpretability and weaker theoretical guarantees compared to explicitly specified generative models.

Several recent articles have contributed important perspectives on graph structure learning. \cite{dong19learning} and \cite{mateos19connecting} provided early overviews of model-based learning with a focus on tools from graph signal processing (GSP), addressing primarily the problem of learning graph topology from node-level observations. \cite{zhu21survey} and \cite{attali24rewiring} surveyed graph structure learning in the GML literature, where the emphasis is on task-driven rewiring of graphs, while \cite{10.1109/TPAMI} and \cite{zhu22survey} focused on deep generative models for graph generation.
This review aims to bridge the various research communities through three main contributions. 
First, 
we provide a unified framework (Fig.~\ref{fig:paradigm}) that places graph topology learning and graph generation, together with their model-based and data-driven formulations, under a common perspective, highlighting both their complementary strengths and their potential synergies. 
Second, the proposed framework naturally suggests new problem formulations that span traditional boundaries.  
For example, rather than treating graph generation and graph signal generation as separate stages, one might think about combining the $h$-map for $G$ and $f$-map for $\mathbf{x}$, thus linking the graph-parameter space $\Theta$ directly to the graph-signal space ${\cal X}$.  
Third, this unifying perspective may also lead to the transfer of methodologies across traditionally separate communities. For example, GSP concepts such as graph filtering and signal smoothness may be combined with or incorporated into the design of deep graph generative models, \dt{while advances in modern generative learning may in turn inspire new approaches to graph topology inference.} 
 

This review article is structured as follows. 
Section \ref{sec:framework} presents the overall framework, formulates the problem instances, and discusses the two families of approaches.  
Subsequently, Sections \ref{sec:graph-topology} and \ref{sec:graph-generation} focus on learning graph topology and learning graph generation, respectively, and Section \ref{sec:downstream-task}  discusses their extensions to downstream tasks. 
Section \ref{sec:bridge} bridges these problem instances, focusing in particular on new problem formulations and the transfer of methodologies across different communities. 
Finally, Section \ref{sec:perspective} concludes with future perspectives on the problem of graph structure learning and beyond.

\section{Framework}
\label{sec:framework}
In this section, we present the unifying framework behind our review. We first introduce the main problem instances, together with  the relevant mathematical notation (Section \ref{sec:problem_form}), and then describe the families of approaches in detail (Section \ref{sec:family_approach}). 
To fix notation, in the rest of this article, each graph, denoted by $G$, consists of a node set $\mathcal{V}$ of $N$ nodes and an edge set $\mathcal{E}$ with edge weights represented by an adjacency matrix $\mathbf{W} \in \mathbb{R}^{N \times N}$. The node observations are represented by a graph signal $\mathbf{x}: \mathcal{V} \rightarrow \mathbb{R}$.  

\subsection{Problem instances} \label{sec:problem_form}

We first define the fundamental instances of the graph structure learning problem that are discussed in this review.  

\begin{problem}[\bf Learning graph topology] \label{prob:topo}
    Given a set of $M$ graph signals $\{\mathbf{x}_i\}_{i=1}^M \in \mathbb{R}^N$, the goal is to learn a graph $G$ of $N$ nodes whose topology is represented by an adjacency matrix $\mathbf{W} \in \mathbb{R}^{N \times N}$, according to a known mapping $f_\mathbf{X}: {\cal G} \to {\cal X}$ (or an unknown $\widehat{f}_\mathbf{X}$) from $G$ to $\mathbf{x}$. 
\end{problem}


\begin{problem}[\bf Learning graph generation] \label{prob:graph}
    Given a set of $M$ graphs $\{G_i\}_{i=1}^M$ whose topologies are represented by adjacency matrices $\{\mathbf{W}_i\}_{i=1}^M \in \mathbb{R}^{N \times N}$ (and each potentially with graph signals $\{\mathbf{x}_i\}_{i=1}^D \in \mathbb{R}^N$), the goal is to learn a generative model, according to a known mapping $h_\mathbf{G}: \Theta \to {\cal G}$ (or an unknown $\widehat{h}_\mathbf{G}$) that represents a probability distribution, such that a new graph instance $G_{i+1}$ with an adjacency matrix $\mathbf{W}_{i+1}$ can be generated. 
\end{problem}


Although Problem \ref{prob:topo} and Problem \ref{prob:graph} have traditionally been studied separately by different communities, they correspond to different stages of the proposed generation process of graph-related data, a connection we will revisit in  Section \ref{sec:bridge}.
{For both problems, it is often desirable to impose topological constraints on the learned or generated graph, such as its node degree distribution. We do not treat this as a separate problem instance, but we discuss constrained graph structure learning whenever relevant.} 
Meanwhile, the above-mentioned formulations do not explicitly consider a downstream task. When such a task is at hand, these formulations may be extended as follows:

\begin{taggedproblem}{3}[\bf Task-driven graph structure learning] \label{prob:topo-task} 
Given a set of $M$ graph signals $\{\mathbf{x}_i\}_{i=1}^M \in \mathbb{R}^N$ (or $M$ graphs $\{ G_i \}_{i=1}^M$) and additional side information pertaining to a downstream task, the goal is to learn a graph $G$ whose topology is represented by an adjacency matrix $\mathbf{W} \in \mathbb{R}^{N \times N}$ (or the parameters of a graph generative model $\theta \in \Theta$). In particular, the selection criterion is influenced by the downstream task which can be defined by labels associated with the given data, or an optimization objective specified by the task.
\end{taggedproblem}

\subsection{Families of approaches} \label{sec:family_approach} 
Next, the existing literature can be broadly categorized into the following two families of approaches: 
\begin{approach}[\bf Model-based learning] \label{app:model} Algorithms belonging to this family assume that $f_\mathbf{X}$ or $h_\mathbf{G}$ is already known. 
Early examples involving  a known 
$f_\mathbf{X}$  include 1) probabilistic graphical models (PGMs) \cite{Koller09}, where one aims to learn the conditional independence structure, represented in the form of a graph, from observations made on random variables which are nodes in the graph, and 2) dynamical processes on graphs, such as network diffusion or disease propagation \cite{Myers2010,GomezRodriguez_2010}. More recently, there is a surge of interest from the GSP community in this problem \cite{dong19learning,mateos19connecting}, where one wishes to learn the topology of the graph that supports the observed signals according to certain prior knowledge that is captured by a known model $f_\mathbf{X}$ (an instance of Problem 1). For example, the smoothness assumption on $\mathbf{x}$ can be interpreted as $\mathbf{x}$ following a multivariate Gaussian distribution (hence a known $f_\mathbf{X}$ as a stochastic map) with the inverse covariance being the corresponding graph Laplacian matrix, and the goal is to learn $G$ from $\mathbf{x}$. Both PGM- and GSP-based approaches emphasize the importance of modeling: the likelihood of the observed data is evaluated given a candidate graph, and learning often amounts to performing a maximum-likelihood (MLE) or maximum-a-posteriori (MAP) estimation.

A complementary line of model-based learning focuses on a known 
$h_\mathbf{G}$ for the entire graph $G$, where each data point is a graph instance. For example, consider a stochastic block model (SBM) (hence a known $h_\mathbf{G}$ as a stochastic map) with parameters $p$ and $q$ for the intra- and inter-community edge probabilities, respectively. The goal is to learn $p$ and $q$ from a collection of observed graphs $\{G_i\}_{i=1}^M$ (an instance of Problem 2), often via an MLE approach \cite{wiuf06likelihood,tabouy20variational}; once learned, $p$ and $q$ can then be used to generate new graph instances.
\end{approach}

\begin{approach}[\bf Data-driven learning] \label{app:data}
In practice, the mapping $f_\mathbf{X}$ or $h_\mathbf{G}$ may not be readily available, and  may therefore need to be learned from data. We first consider an unknown $\widehat{f}_\mathbf{X}$ for $\mathbf{x}$. In this setting, one may rely on a prototype model with unknown parameters that determine its characteristics, or equip the prototype model with learnable components, thereby offering more flexibility. 
In this case, one may observe pairs of node-level observations and graph topologies, from which a mapping from $G$ to $\mathbf{x}$ can be learned and later used to infer a graph from new node signals (an instance of Problem 1). As a concrete example, we can think of $\mathbf{x}$ being generated from a fixed but unknown graph filter (e.g., polynomial filter or diffusion kernel) applied to random noise, or from an unknown diffusion process on $G$. The goal is to learn the filter or diffusion process $\widehat{f}_\mathbf{X}$ from observed pairs of $\mathbf{x}$ and $G$, which serve as training samples. The learned mapping can then be used to infer $G$ given new observations $\mathbf{x}$. 

Similarly, we may consider an unknown mapping $\widehat{h}_\mathbf{G}$ for $G$. In this case, the training data may consist of graph-level observations, such as graph instances collected over time. These graphs may be viewed as realizations of an unknown probabilistic model $\widehat{h}_\mathbf{G}$. The goal is to learn $\widehat{h}_\mathbf{G}$ from a collection of observed $\{G_i\}_{i=1}^M$ (an instance of Problem 2); once learned, $\widehat{h}_\mathbf{G}$ can then be used to generate new graph instances. Representative works include deep graph generative models \cite{10.1109/TPAMI,zhu22survey}.
\end{approach}

Problems \ref{prob:topo} and \ref{prob:graph} correspond to the second and first columns of Fig.~\ref{fig:paradigm}, respectively, while Approaches \ref{app:model} and \ref{app:data} correspond to the first and second rows, respectively. Problem \ref{prob:topo-task} can be considered as an extension of both Problems \ref{prob:topo} and \ref{prob:graph} by incorporating downstream tasks. The following sections discuss each of these problems in greater detail. 


\section{Learning Graph Topology}
\label{sec:graph-topology}

\subsection{Learning graph topology with a known mapping $f_{\bf X}$}
\label{sec:known-fX}


We first consider Problem \ref{prob:topo} where a set of $M$ graph signals $\{\mathbf{x}_i\}_{i=1}^M \in \mathbb{R}^N$ is observed according to a known mapping $f_\mathbf{X}: {\cal G} \to {\cal X}$ from $G$ to $\mathbf{x}$. Given  this mapping, the goal is to learn from $\mathbf{x}$ a graph $G$ of $N$ nodes represented by an adjacency matrix $\mathbf{W} \in \mathbb{R}^{N \times N}$ (or an equivalent representation). We discuss several representative examples of this problem and its extensions. 


\begin{example}[\bf Smooth signals on graphs] \label{example:iiia-smooth}
Consider a signal $\mathbf{x}$ that is smooth on the graph $G$, where smoothness is intuitively understood as neighboring nodes in the graph having similar signal values or distributions. From a probabilistic viewpoint, this is often encoded by a known mapping $f_\mathbf{X}$ which represents a distribution of $\mathbf{x}$ that is characterized by a representation of $G$.
\end{example}

Smoothness is one of the most natural assumptions on the characteristics of the graph signal. Early instances of Example \ref{example:iiia-smooth} include those characterized by the PGMs \cite{Koller09}. In PGMs, observations are made on a collection of random variables in the form of a matrix $\mathbf{X} \in \mathbb{R}^{N \times M}$, where the rows correspond to the random variables, the existence of conditional independence among which is captured by the presence of edges in a graph representation (hence an implicit smoothness assumption). In the case of Gaussian Markov random fields (GMRFs), this leads to a joint distribution encoded by $f_\mathbf{X}$: 
\begin{equation}
f_\mathbf{X}: p(\mathbf{x}|\boldsymbol{\Theta}) = \frac{|\boldsymbol{\Theta}|^{1/2}}{(2 \pi)^{N/2}} \text{exp} \big( -\frac{1}{2} \mathbf{x}^T \boldsymbol{\Theta} \mathbf{x} \big),
\label{eq:gmrf}
\end{equation}
where $\boldsymbol{\Theta} \in^{N \times N} $ is the inverse covariance or precision matrix. In this case, learning the graph $G$ amounts to learning $\boldsymbol{\Theta}$, often via an MLE approach. For example, the graphical lasso method proposes to solve the following problem \cite{Friedman08}: 
\begin{equation}
\max_{\boldsymbol{\Theta}}~\text{logdet}(\boldsymbol{\Theta}) - \mathrm{tr}(\hat{\boldsymbol{\Sigma}} \boldsymbol{\Theta}) - \rho ||\boldsymbol{\Theta}||_1,
\label{eq:glasso}
\end{equation}
where $\hat{\boldsymbol{\Sigma}} = \frac{1}{M-1} \mathbf{X} \mathbf{X}^T$ is the empirical covariance, $\text{logdet}(\cdot)$ and $\mathrm{tr}(\cdot)$ represent the log determinant and trace operators, respectively, and $\rho$ is a regularization parameter. 

More recently, the field of GSP has made advances in solving Example \ref{example:iiia-smooth} by making an explicit assumption on signal smoothness. Consider the graph Laplacian matrix and its eigendecomposition, i.e., $\mathbf{L} = \mathbf{U} \mathbf{\Lambda} \mathbf{U}^T$. Under a factor analysis model of a graph signal $\mathbf{x}$ (columns of $\mathbf{X}$ above), i.e., $\mathbf{x} = \mathbf{U} \mathbf{h} + \mathbf{n}$, where $\mathbf{h} \sim \mathcal{N}(\mathbf{0},\mathbf{\Lambda}^\dagger)$ represents a coefficient vector under the eigenvector basis $\mathbf{U}$ and $\mathbf{n} \sim \mathcal{N}(\mathbf{0}, \sigma_N^2 \mathbf{I})$ is additive Gaussian noise, the distribution of $\mathbf{x}$ follows \cite{Dong16}:
\begin{equation}
f_\mathbf{X}: p(\mathbf{x}|\mathbf{L}) \sim \mathcal{N}(\mathbf{0}, \mathbf{L}^\dagger + \sigma_N^2 \mathbf{I}), 
\label{eq:smooth}
\end{equation}
where $\dagger$ represents the Moore-Penrose pseudo-inverse. Based on this, \cite{Dong16} proposes to solve for $\mathbf{L}$ (equivalently the graph adjacency matrix) via an optimization problem whose objective includes a trace term of the so-called Laplacian quadratic form, i.e., $\mathrm{tr}(\mathbf{X}^T \mathbf{L} \mathbf{X})$, which explicitly penalizes variation of signal values across connected nodes, hence promoting smoothness on the learned graph. 
Other approaches based on a similar objective include \cite{Kalofolias16,Egilmez17,Chepuri17}. Going beyond the Gaussian assumption, recent work has also considered learning from signals corrupted with exponential family noise \cite{shi25graph}. 

We note that there is a clear link between the PGM and GSP approaches to Example \ref{example:iiia-smooth}; indeed, if we restrict the precision matrix $\boldsymbol{\Theta}$ to be a symmetric and positive definite matrix with non-positive off-diagonal entries, of which the graph Laplacian $\mathbf{L}$ is an example, the ways in which signal smoothness is being promoted in the two approaches are equivalent (i.e., through the minimization of the Laplacian quadratic form). 
More recent work has extended this notion of smoothness by taking into account additional node attributes that, together with the graph topology, influence the observed graph signals. In \cite{pu21kernel}, this leads to the following distribution of $\mathbf{x}$: 
\begin{equation}
f_\mathbf{X}: \mathbf{x} \sim \mathcal{N}(\mathbf{0}, \mathbf{L}^\dagger \otimes \mathbf{K} +  \sigma_N^2 \mathbf{I}),
\label{eq:kernel-smooth}
\end{equation}
where $\mathbf{K}$ is a chosen kernel matrix defined over pairs of node attributes, and $\otimes$ represents the Kronecker product. This in turn leads to an optimization problem over $\mathbf{L}$ with a regularization term of $\mathbf{x}^T (\mathbf{L} \otimes \mathbf{K}^{\dagger}) \mathbf{x}$. Interestingly, if one associates the kernel matrix with another graph Laplacian $\mathbf{L}_{\mathbf{K}} = \mathbf{K}^{\dagger}$ and defines the Kronecker product $\mathbf{L}_{\otimes} = \mathbf{L} \otimes \mathbf{L}_{\mathbf{K}}$ as a Laplacian-like operator, then this term becomes $\mathbf{x}^T \mathbf{L}_{\otimes} \mathbf{x}$, which can be interpreted as a modified Laplacian quadratic form that promotes signal smoothness on the learned graph (represented by $\mathbf{L}$) and in the chosen kernel space (represented by $\mathbf{K}$) simultaneously. 
On the other hand, a few approaches have made the smoothness assumption not on the graph signal itself, but a constituent of it. For example, \cite{guo23graph} posits that the signal is a combination of a smooth component and a remainder explained by non-graph-related regressors; \cite{watanabe23learning} and \cite{ma23gist} both consider signals that can be expressed as a linear combination of building block patterns, which are then assumed to be smooth on the learned graph. 

The additional constraint implicitly imposed in the kernel space of node attributes makes the approach in \cite{pu21kernel} robust against randomly missing entries in the node observations, a scenario also considered in \cite{berger20efficient} and \cite{karaaslanli21graph}. In these approaches, a mask matrix that indicates the missing entries is assumed known, and an optimization problem with a similar smoothness-based objective as above is solved. A similar imperfect setting is considered in \cite{buciulea22learning} and \cite{nguyen26learning}, where observations are only made on a subset of nodes in the graph and a subgraph among those is learned. In \cite{buciulea22learning}, this is achieved by decomposing the Laplacian quadratic form into three terms that correspond to the observed nodes, unobserved nodes, and interactions between them, which are learned in a joint optimization problem. 

In parallel, an increasingly rich literature considers different forms of learning output.  In many cases, the smoothness measure is kept in the objective of the optimization problem, with additional terms constraining the output form. 
One output form of particular interest is the simultaneous learning of multiple graphs and, as a typical example, time-varying graphs. The early works of \cite{hallac17network} and \cite{kalofolias17learning} generalize the classical graphical lasso and GSP-based topology learning, respectively, to the time-varying setting, where the learned graphs at consecutive timestamps are constrained to be similar. While \cite{kalofolias17learning} enforces this via the Frobenius norm of differences in the weighted adjacency matrices, \cite{yamada19time} further constrains the number of edges modified across consecutive timestamps to be small via an $\ell_1$-norm based regularization. Various extensions have since been proposed to deal with online streaming data \cite{natali22learning}, partial node observations \cite{yang20network,bagheri24joint}, and the presence of hidden nodes \cite{ye24time}. Going beyond time-varying graphs, \cite{danaher14joint} and \cite{navarro22joint} generalize the graphical lasso and GSP approaches, respectively, in learning multiple graphs, while \cite{karaaslanli25multiview} proposes to learn a multi-view graph together with a consensus topology, given node observations from each view. Similarly, \cite{saboksayra21online} associates each view with a class label in a classification task, therefore learning class-specific adjacency matrices. 

As a final remark, we can see above that algorithmically the smoothness criterion is often incorporated into an optimization problem, which can then be solved by for example the block-coordinate descent (BCD) or alternating direction method of multipliers (ADMM) algorithms. In practice, it may be challenging to tune the hyperparameters in these formulations, such as $\rho$ in Eq.~(\ref{eq:glasso}). To address this, \cite{shrivastava20glad} proposes to unroll an algorithm that solves a reformulated version of Eq.~(\ref{eq:glasso}), where $\rho$ is modeled as a problem-dependent neural network. Similarly, \cite{pu21learning} proposes to unroll an algorithm designed to solve a GSP-based formulation, and \cite{wasserman24graph} further considers a Bayesian extension that allows for the estimation of uncertainty of learned edges. All these can thus be considered as a form of algorithm learning, 
although they are still based on prototypes of formulations that explicitly promote signal smoothness.

\begin{example}[\bf Dynamical processes on graphs] \label{example:iiia-dynamical}
Consider a signal that is the outcome of a physical dynamical process on the graph, such as graph-based diffusion. This process is encoded by a known mapping $f_\mathbf{X}$ which maps an initial state of the signal and a graph $G$ to observed signals $\mathbf{x}$. 
\end{example}

Such a process naturally encodes the temporal dynamics behind the observed signals on the graph and, as a consequence, the observations often come with a temporal dimension. 
Early instances of graph diffusion can be found in epidemiology, where a disease is assumed to have been transmitted over a contact network among a group of individuals, having originated from a random individual in the network. The transmission probability from node $i$ to node $j$, given that $i$ is already infected, is governed by a mapping $f_\mathbf{X}$, which is often a function of the graph topology. For example, in \cite{Myers2010}, this probability is defined directly using the edge weight $w_{ij}$:
\begin{equation}
f_\mathbf{X}: p\big(x_j \big| x_i, w_{ij}\big) = h\big(x_j - x_i\big) w_{ij}, 
\end{equation}
where $x_i$ and $x_j$ are the infection times of nodes $i$ and $j$, respectively, with $x_i < x_j$, and $h(t)$ is a transmission time model that is assumed known. Given only observations on the node states (such as $x_i$ for all nodes), the goal is to learn the topology of the unknown contact network $G$. In the case of \cite{Myers2010}, learning $G$ amounts to learning the edge weights $w_{ij}$ as the conditional probabilities of disease transmission, which can be done via an MLE approach. A similar approach based on a binary graph topology and a uniform transmission probability $w_{ij}$ can be found in \cite{GomezRodriguez_2010}. 

Going beyond infectious disease models, the vector autoregressive model (VARM) and its variants have been used to study temporal dynamics on graphs, which introduce both temporal and spatial dependencies in the data. For example, in \cite{Mei17}, the signal at time step $t$, $\mathbf{x}[t]$, is represented as a linear combination of transformed observations in the past $K$ time steps and a random noise process $\mathbf{n}[t]$:
\begin{equation}
f_\mathbf{X}: \mathbf{x}[t] = \sum_{k=1}^K p_k(\mathbf{W}) \mathbf{x}[t-k] + \mathbf{n}[t],
\label{eq:var}
\end{equation}
\noindent where $p_k(\mathbf{W})$ is a polynomial in the adjacency matrix $\mathbf{W}$. Given temporal observations $\mathbf{x}[t]$, \cite{Mei17} proposes to solve an optimization problem over $\mathbf{W}$ by minimizing the approximation error according to Eq.~(\ref{eq:var}), together with a sparsity constraint on both $\mathbf{W}$ and the polynomial coefficients. Alternatively, the same signal model can be posed as a Gaussian process with the choice of the kernel function encoding the spatiotemporal dependencies \cite{cui24gaussian}. A VARM can also be interpreted as a state-based model (similar to a Kalman filter). For example, \cite{liu19graph} posits the following signal model:
\begin{equation}
f_\mathbf{X}: \mathbf{x}[t] = \mathbf{U} \mathbf{h}[t] + \mathbf{n}[t],~\mathbf{h}[t] = \mathbf{A} \mathbf{h}[t-1] + \mathbf{v}[t],
\label{eq:state1}
\end{equation}
where $\mathbf{U}$ is chosen to be the eigenvector matrix of $\mathbf{L}$ (hence it can be interpreted as a graph Fourier basis as in \cite{Dong16}), and $\mathbf{h}$ is a latent state variable governed by the state transition matrix $\mathbf{A}$. It is further assumed that $\mathbf{A} = \mathbf{U}^T \mathbf{R} \mathbf{U}$, where $\mathbf{R}$ is a diagonal matrix capturing correlation in time, and $\mathbf{n}[t] \sim \mathcal{N}(\mathbf{0}, \sigma_N^2 \mathbf{I})$ and $\mathbf{v}[t] \sim \mathcal{N}(\mathbf{0},\mathbf{\Lambda}^\dagger)$, where $\mathbf{\Lambda}$ is the eigenvalue matrix of $\mathbf{L}$. Under these assumptions, the temporal difference signal admits the following distribution: 
\begin{equation}
p(\mathbf{d}|\mathbf{L}) \sim \mathcal{N} \Big( \mathbf{0}, \big( \mathbf{L}^\dagger + \sigma_N^2 (\mathbf{I} + \mathbf{R} \mathbf{R}^T) \big) \Big). 
\label{eq:state2}
\end{equation}
Therefore, it can be seen that the difference signal $\mathbf{d}$ is smooth on the graph, and the model in Eq.~(\ref{eq:state2}) encodes a notion of spatiotemporal smoothness governed by $\mathbf{A}$, which introduces a space-time coupling. Based on this, \cite{liu19graph} formulates an optimization problem to learn $\mathbf{L}$ and $\mathbf{R}$ jointly. A similar model, based on the graphical lasso formulation and also considering partial node observations, can be found in \cite{javaheri24learning}. 
More generally, the dynamical process can incorporate additional influence from instantaneous effects, by combining a VARM with a structural equation model (SEM) into a structural vector autoregressive model (SVARM) \cite{shen19nonlinear}, or by extending the state-based interpretation above into a state-space model \cite{coutino20state}.

Apart from those based on the graph Laplacian $\mathbf{L}$, approaches discussed above can generally be used to learn a graph with directed edges (i.e., an asymmetric adjacency matrix $\mathbf{W}$). This ability to learn directed graphs is an important difference from the approaches to Example \ref{example:iiia-smooth}. Of particular interest are directed acyclic graphs (DAGs), which come with an interpretation of causal relations between the variables that are represented by the nodes in the graph. The reader is referred to \cite{mateos26concomitant} for a historical overview of DAG learning. The literature mostly adopts the linear SEM (while the discussion here focuses on learning DAGs, the SEM has also been used in learning generic directed graphs \cite{ceci20graph}): 
\begin{equation}
f_\mathbf{X}: \mathbf{x} = \mathbf{W}^T \mathbf{x} + \mathbf{n},
\label{eq:sem}
\end{equation}
where $\mathbf{W}$ represents the causal dependencies between the nodes (i.e., the $ij$-th entry of $\mathbf{W}$, $w_{ij}$, is nonzero if there is a causal link from $i$ to $j$), and $\mathbf{n}$ is observational noise. Under this model, existing approaches mainly differ in how the acyclicity constraint on $\mathbf{W}$ is being enforced. While traditional approaches mostly rely on a combinatorial constraint,  NOTEARS \cite{zheng18dags} proposes to enforce acyclicity via the constraint $h(\mathbf{W}) = \mathrm{tr}(e^{\mathbf{W} \circ \mathbf{W}}) - n = 0$, where $\circ$ represents the Hadamard product and $e^{(\cdot)}$ the matrix exponential, thus turning a discrete optimization into a continuous equality-constrained program which is then solved via the augmented Lagrangian method. Various extensions have been proposed since then: \cite{pamfil20dynotears} and \cite{dacunto23multiscale} extend \cite{zheng18dags} to deal with time-series data, \cite{ng20sparsity} analyzes the role of sparsity in $\mathbf{W}$, \cite{misiakos23learning} further considers sparsity in $\mathbf{n}$ which is interpreted as having a few root causes of the observations, and \cite{saboksayr24colide} proposes a joint estimation of $\mathbf{W}$ and the unknown noise level in $\mathbf{n}$. 

More recently, the literature has studied an interesting example of network processes often encountered in social and economic sciences, i.e., games played on networks \cite{jackson2014games}. These are strategic decision-making processes where each individual in the network takes an action to maximize their own payoff, which in turn depends on the actions of their neighbors. For example, in a network game with linear-quadratic payoffs \cite{ballester2006who}, an individual $i$ chooses their action $x_i$ to maximize their payoff $u_i$:
\begin{equation}
u_i = b_i x_i -\frac{1}{2}x_i^2 + \beta x_i \sum_{j \sim i} w_{ij} x_j, 
\label{eq:utility}
\end{equation}
where $b_i$ is the marginal benefit $i$ receives by taking the action $x_i$. It is shown in \cite{ballester2006who} that given certain assumptions, this game admits the following unique and stable Nash equilibrium: 
\begin{equation}
f_\mathbf{X}: \mathbf{x} = \left( \mathbf{I} - \beta \mathbf{W} \right)^{-1}\mathbf{b}, 
\label{eq:equi}
\end{equation}
where $\beta$ is a parameter that determines the nature of the strategic interaction. \cite{leng20learning} exploits this relationship to design an optimization problem that learns an unknown $\mathbf{W}$ and parameter $\beta$ from observed equilibrium actions $\mathbf{x}$. 
Another example similar to network games is the consensus process studied in \cite{zhu20network}. 

As a remark, many of the dynamical processes above can be summarized in the following generic form proposed in \cite{wai19joint}:
\begin{equation}
f_\mathbf{X}: \dot{x}_i(t) = g_i \big( x_i(t) \big) + \sum_{j \sim i} w_{ij} h \big( x_i(t), x_j(t); \boldsymbol{\theta}_i \big),
\label{eq:wai19}
\end{equation}
where $x_i(t)$ and $\dot{x}_i(t)$ are the time-series observation at node $i$ and its time derivative at time $t$, respectively, $g_i$ is a self-update function on $i$, and $h$ (with its parameters $\boldsymbol{\theta}_i$) governs the potentially nonlinear dynamical process on the graph and its impact on $i$. As examples, \cite{wai19joint} specializes this to gene dynamics and opinion dynamics based on two different choices of $h$. To learn the edge weights $w_{ij}$, it assumes the existence of stationary points of the process above, and designs an inverse problem that optimizes over $w_{ij}$ and the parameters $\boldsymbol{\theta}_i$ given perturbed stationary points.

\begin{example}[\bf Filtering processes on graphs] \label{example:iiia-filter}
Consider a signal that is the outcome of a filtering process on the graph. The filter characteristic is encoded by a known mapping $f_\mathbf{X}$ which maps an initial graph signal $\mathbf{x}_0$ and a graph $G$ to observed signals $\mathbf{x}$. 
\end{example}

The smoothness-based model and some of the dynamical processes discussed above, e.g., Eq.~(\ref{eq:var}), can be unified under the language of GSP via a filtering process. In this case, the observed signals $\mathbf{x}$ can be written as: 
\begin{equation}
f_\mathbf{X}: \mathbf{x} = g(\mathbf{S})\mathbf{x}_0 = \mathbf{U} g(\mathbf{\Lambda}) \mathbf{U}^T \mathbf{x}_0, 
\label{eq:filter}
\end{equation}
where $\mathbf{S} = \mathbf{U} \mathbf{\Lambda} \mathbf{U}^T$ is a (symmetric) graph shift operator representing the topology of graph $G$, and $g(\cdot)$ defines the characteristics of the filter via the interpretation of $\mathbf{U}^T \mathbf{x}_0$ being a Fourier-like transform on the graph. This signal model is more general and flexible than the ones above: indeed, if $S$ is chosen to be the graph Laplacian $\mathbf{L}$ and $g(\cdot)$ a low-pass filter \cite{ramakrishna20user}, this corresponds to the smoothness-based model in Eq.~(\ref{eq:smooth}); alternatively, $g(\cdot)$ may act as a frequency-selective filter that only allows specific graph frequencies to pass, resulting in a band-limited (but not necessarily smooth) signal \cite{batreddy21graph}. On the other hand, the VARM can also be interpreted as graph filtering with $\mathbf{S}$ being the adjacency matrix. 

Broadly speaking, two main approaches have been deployed in the context of graph topology learning. The first posits that the graph filter corresponds to a diffusion-like process on the graph, such as that in \cite{Mei17} discussed above. Other notable processes studied in the literature include heat diffusion in \cite{Thanou17} and the consensus dynamics in \cite{zhu20network}. For example, in \cite{Thanou17},  $g(\cdot)$ represents the heat kernel based on the choice of $\mathbf{S}$ being the graph Laplacian $\mathbf{L}$: 
\begin{equation}
f_\mathbf{X}: \mathbf{x} = g(\mathbf{L}) \mathbf{x}_0 = e^{-\tau \mathbf{L}}\mathbf{x}_0, 
\end{equation} 
where $\tau$ is a parameter that controls the speed of diffusion. \cite{Thanou17} further assumes that the signal is a linear combination of the outcomes of a small number of heat diffusion processes  (hence similar in spirit to \cite{misiakos23learning}), which may have originated from different nodes with different parameters $\tau$. A dictionary-learning algorithm is then applied to solve for $\mathbf{L}$ given $\mathbf{x}$. 

In the second approach, $g(\cdot)$ represents a generic degree-$K$ polynomial graph filter: 
\begin{equation}
f_\mathbf{X}: \mathbf{x} = g(\mathbf{S}) \mathbf{x}_0 = \sum_{k = 0}^{K}g_k \mathbf{S}^k \mathbf{x}_0, 
\label{eq:filter2}
\end{equation}
where $\mathbf{S}$ can be chosen as either the adjacency matrix \cite{Segarra17a} or graph Laplacian \cite{Pasdeloup18,zhu20network}. Under the assumption of a symmetric $\mathbf{S}$ and that the initial signal $\mathbf{x}_0$ is i.i.d. Gaussian, this has the interpretation of a stationary process on the graph \cite{marques17stationary}. In this case, the covariance matrix $\mathbf{C} = \mathbb{E}(\mathbf{x} \mathbf{x}^T)$ commutes with $\mathbf{S}$; hence, the two share the same set of eigenvectors. Once the eigenvectors of $\mathbf{S}$, which are termed spectral templates in \cite{Segarra17a}, are obtained from $\mathbf{C}$, the eigenvalues are found via an optimization problem, often together with a sparsity constraint on $\mathbf{S}$. 

Similar to Example \ref{example:iiia-smooth} and Example \ref{example:iiia-dynamical}, the standard approaches above have been extended to a variety of settings. In terms of the input information, \cite{buciulea22learning} considers the partial node observation setting with the presence of hidden nodes. Regarding the form of output, \cite{einizade23learning} and \cite{zhang23product} consider the task of learning product graphs. For example, in \cite{zhang23product}, the graph signal admits the following distribution: 
\begin{equation}
f_\mathbf{X}: \mathbf{x} = \big[ \sum_{p=0}^{K_P} \sum_{q=0}^{K_Q} g_{pq} (\mathbf{W}_P)^p \otimes (\mathbf{W}_Q)^q \big] \mathbf{x}_0 + \mathbf{n}, 
\end{equation}
where $\mathbf{W}_P$ and $\mathbf{W}_Q$ are the adjacency matrices for the graphs $G_P$ and $G_Q$ that form the product graph, $\{g_{pq}\}$ are the filter coefficients, and $\mathbf{n}$ is the noise. The approach based on spectral templates can then be applied to learn the graph topologies of $G_P$ and $G_Q$ separately, before they are combined to reconstruct $G$. Another example is \cite{batreddy25learning}, which learns bipartite graphs using the same approach. Finally, there are also efforts that aim to deal with the time-varying setting \cite{shafipour20online} and multi-graph setting \cite{navarro24joint,zhang24learning}.

\subsection{Learning graph topology with an unknown mapping $\widehat{f}_\mathbf{X}$}
\label{sec:unknown-fX}


We now consider the data-driven counterpart of Section \ref{sec:known-fX} in solving Problem \ref{prob:topo}. In this case, the mapping from the graph $G$ to the observed signals $\mathbf{x}$ is either completely unknown, or follows a prototype mapping but with unknown parameters that determine its precise characteristics. The goal is to learn from $\mathbf{x}$ the graph $G$ by learning the unknown mapping $\widehat{f}_\mathbf{X}$, via training examples and a criterion on the quality of the learned mapping. There are two key considerations in this case: 1) how the relation between $G$ and $\mathbf{x}$ is relaxed from a known to an unknown mapping, and 2) how this relaxation impacts the corresponding learning strategy. 
In what follows, we discuss how the various approaches differ in these two aspects, by focusing on the extension of two examples in Section \ref{sec:known-fX}.

\begin{example}[\bf Learning unknown filtering processes] \label{example:iva-filter}
Consider a signal that is the outcome of a filtering process on the graph. The filter characteristic is encoded by an unknown mapping $\widehat{f}_\mathbf{X}$ which maps an initial graph signal $\mathbf{x}_0$ and a graph $G$ to observed signals $\mathbf{x}$.
\end{example}

We begin with Example \ref{example:iva-filter} as it provides a smoother transition in terms of the methodologies covered in this section. In the graph filtering formulation, the signal model generally follows that in Eq.~(\ref{eq:filter2}), where $\{g_k\}$ are the filter coefficients. Note that most approaches discussed after Eq.~(\ref{eq:filter2}) assume a stationary graph process, in which case the eigenvectors of the graph shift operator $\mathbf{S}$ can be obtained directly from that of the covariance of the observed signals, and the learning of $\mathbf{S}$ can be achieved without the knowledge of $\{g_k\}$. When this assumption does not hold for generic graph signals, however, learning $\{g_k\}$, i.e., what constitutes the unknown mapping $\widehat{f}_\mathbf{X}$, often becomes necessary, either explicitly or implicitly via learning $g(\mathbf{S})$ directly. 

We first discuss an interesting case that bridges the graphical lasso and GSP formulations. Recall the GMRF in Eq.~(\ref{eq:gmrf}), where the support of the precision matrix $\boldsymbol{\Theta}$ indicates the presence of conditional dependence between the random variables represented by the nodes in the graph. From this perspective, learning $\boldsymbol{\Theta}$ can be interpreted as learning a graph, and in the case of $\boldsymbol{\Theta}$ being the graph Laplacian it leads to the smoothness model in Eq.~(\ref{eq:smooth}). In \cite{buciulea25polynomial}, this specific case is relaxed such that $\boldsymbol{\Theta} = g(\mathbf{S}) = \sum_{k = 0}^{K}g_k \mathbf{S}^k$ where $g(\cdot)$ is a generic polynomial. In other words, $\mathbf{x}$ still follows the same distribution as in Eq.~(\ref{eq:gmrf}), but its interpretation now depends on the characteristics of $g(\cdot)$: if it is no longer a monotonically increasing function of the eigenvalue of $\mathbf{S}$, the model will not necessarily promote signal smoothness but capture a more flexible behavior. Given that $\boldsymbol{\Theta}$ and $\mathbf{S}$ commute, \cite{buciulea25polynomial} proposes to solve the following joint optimization problem: 
\begin{equation}
\max_{\boldsymbol{\Theta},\mathbf{S}}~\text{logdet}(\boldsymbol{\Theta}) - \mathrm{tr}(\hat{\boldsymbol{\Sigma}} \boldsymbol{\Theta}) - \rho ||\boldsymbol{\Theta}||_0,~~\text{s.t.}~~\boldsymbol{\Theta} \mathbf{S} = \mathbf{S} \boldsymbol{\Theta}, 
\label{eq:pglasso}
\end{equation}
where $\{g_k\}$ are implicitly learned via $\boldsymbol{\Theta}$. A similar approach is proposed in \cite{egilmez19graph}, where $\boldsymbol{\Theta} = h_{\theta}(\mathbf{L})^\dagger$ follows a number of specific parametric forms, and hence the associated parameter $\theta$ is explicitly learned together with the graph Laplacian $\mathbf{L}$. 

Moving on to the GSP formulation of Eq.~(\ref{eq:filter2}), where we write $\mathbf{G} = g(\mathbf{S})$ in short, \cite{shafipour18directed} proposes a two-step approach to joint graph topology learning and filter identification. First, given that $\mathbf{x} = \mathbf{G} \mathbf{x}_0$, we have $\boldsymbol{\Sigma}_\mathbf{x} = \mathbf{G} \boldsymbol{\Sigma}_{\mathbf{x}_0} \mathbf{G}^T$, where $\boldsymbol{\Sigma}_{\mathbf{x}_0}$ and $\boldsymbol{\Sigma}_\mathbf{x}$ are the true input and output covariance matrices, respectively. Assuming the former is known, $\mathbf{G}$ can be estimated given the empirical output covariance $\hat{\boldsymbol{\Sigma}}_\mathbf{x} = \frac{1}{m-1} \mathbf{X} \mathbf{X}^T$ by solving a manifold-constrained least-squares problem. Second, once the estimated $\hat{\mathbf{G}}$ is obtained, $\mathbf{S}$ can be learned by finding a sparse matrix that commutes with $\hat{\mathbf{G}}$. This approach based on the constraint on commutativity also helps avoid the need for eigendecomposition and the assumption of a symmetric $\mathbf{S}$ in \cite{Segarra17a} and \cite{Pasdeloup18}. In practice, if the true input covariance is not known, one could instead resort to the knowledge of input-output pairs of an unknown filtering system. For example, \cite{natali20topology} proposes to solve the following joint optimization problem: 
\begin{equation}
\min_{\mathbf{S},\mathbf{g}}~|| \mathbf{X} - \sum_{k = 0}^{K}g_k \mathbf{S}^k \mathbf{X}_0 ||~~\text{s.t.}~~\text{supp}(\mathbf{S}) \subseteq \mathcal{S}, 
\label{eq:filter-id}
\end{equation}
where $\mathbf{X}_0$ and $\mathbf{X}$ are the input and output signals in matrix format, respectively, $\mathbf{g}$ is a vector of filter coefficients $\{g_k\}$, and $\mathcal{S}$ is the set with the support of $\mathbf{S}$ that is assumed known. A similar approach can be found in \cite{rey23robust} which, instead of learning the filter coefficients explicitly, proposes to learn the filter matrix $\mathbf{G} = g(\mathbf{S})$ with a commutativity constraint as in \cite{shafipour18directed}. 

Most approaches discussed in this section and Section \ref{sec:known-fX} follow a deterministic setting, where only a point estimate of the graph topology is obtained. 
To address this, \cite{sevilla24bayesian} proposes a Bayesian approach and considers the following MAP estimation: 
\begin{equation}
\max_{\mathbf{W},\boldsymbol{\theta}}~p(\mathbf{W},\boldsymbol{\theta} | \mathbf{X}_0,\mathbf{X})~~\text{or equivalently}~~\max_{\mathbf{W},\boldsymbol{\theta}}~p(\mathbf{X} | \mathbf{W},\boldsymbol{\theta},\mathbf{X}_0) p(\mathbf{W}) 
\label{eq:filter-map}
\end{equation}
based on the assumptions on the input $\mathbf{X}_0$ and filter coefficients $\boldsymbol{\theta}$. In Eq.~(\ref{eq:filter-map}), the likelihood is governed by the unknown graph filtering process, while the prior represents any knowledge on the graph topology. Given that expressions for both may not be tractable in general, \cite{sevilla24bayesian} proposes drawing a sample from the posterior distribution to approximate the MAP estimate. The algorithm proceeds iteratively: given an estimate of $\boldsymbol{\theta}$, $\mathbf{W}$ is updated via the application of annealed Langevin dynamics; $\boldsymbol{\theta}$ is then updated via gradient descent, and the process repeats. Based on the filtering process in Eq.~(\ref{eq:filter2}), the likelihood is computed using input-output signal pairs together with estimates of $\mathbf{W}$ and $\boldsymbol{\theta}$ (the latter of which is equivalent to $\{g_k\}$). The prior is approximated via a graph neural network (GNN), which is trained using a set of ground truth adjacency matrices. 

\begin{example}[\bf Learning unknown dynamics] \label{example:iva-dynamical}
Consider a signal that is the outcome of a dynamical process on the graph. This process is encoded by an unknown mapping $\widehat{f}_\mathbf{X}$ which maps an initial state of the signal and a graph $G$ to observed signals $\mathbf{x}$. 
\end{example}

We now move on to unknown dynamical processes on graphs. Although many of these processes can be interpreted as graph-based filtering, the methodologies developed in the literature cover a broader range of techniques and increasingly favor a neural network based design, as we shall see in this section. 

As discussed in Section \ref{sec:known-fX}, the classical SEM in Eq.~(\ref{eq:sem}) has been widely used to capture (causal) relations between nodes in a DAG. However, it is an inherently linear model and therefore might be restrictive when nonlinear relations are present. The literature has taken a number of approaches to address this. In \cite{zheng20learning}, the following nonlinear form is considered: 
\begin{equation}
\widehat{f}_\mathbf{X}: x_j = f_j(\mathbf{x}) + n_j, 
\label{eq:dag-mlp}
\end{equation}
where $f_j$ is a generic nonlinear function that governs node $j$'s observation $x_j$, and $n_j$ is the noise. \cite{zheng20learning} chooses $f_j$ to be a multilayer perceptron (MLP) with $L$ hidden layers: $\text{MLP}(\mathbf{x}; \mathbf{A}_j^{(1)}, \cdots, \mathbf{A}_j^{(L)}) = \sigma \Big( \mathbf{A}_j^{(L)} \sigma \big( \cdots \sigma (\mathbf{A}_j^{(1)} \mathbf{x}) \big) \Big)$, where $\mathbf{A}_j^{(\ell)} \in \mathbb{R}^{m_\ell \times m_{\ell-1}}$, $m_0 = N$, and $\sigma$ is a nonlinear activation function. Under this setting, one can see that $x_j$ is independent of $x_i$ if the $i$-th column of $\mathbf{A}_j^{(1)}$ is all zero. This motivates the definition of the edge weights $w_{ij} = ||i\text{th-column}(\mathbf{A}_j^{(1)})||_2$ in a graph of functional dependence. 
Sparsity and the same DAG constraint as in \cite{zheng18dags} can then be imposed on its adjacency matrix $\mathbf{W}$. 
Based on this, \cite{zheng20learning} proposes to solve an optimization problem with respect to $\mathbf{A}_j^{(\ell)}$ by minimizing a reconstruction loss, hence recovering a DAG underlying the data observations. 

An alternative way of incorporating nonlinear relations into the SEM is proposed in \cite{yu19daggnn}. Eq.~(\ref{eq:sem}) can be rewritten in matrix form as $\mathbf{X} = (\mathbf{I}-\mathbf{W}^T)^{-1} \mathbf{N}$, motivating the following nonlinear extension of the SEM: 
\begin{equation}
\widehat{f}_\mathbf{X}: \mathbf{X} = f_2\big[(\mathbf{I}-\mathbf{W}^T)^{-1} f_1(\mathbf{N})\big],
\label{eq:dag-gnn}
\end{equation}
where $f_1$ and $f_2$ are generic functions. In \cite{yu19daggnn}, $f_1$ is chosen to be an identity mapping and $f_2$ a 2-layer MLP: $\text{MLP}\big( (\mathbf{I}-\mathbf{W}^T)^{-1} \mathbf{N}, \mathbf{A}^{(1)}, \mathbf{A}^{(2)} \big) = \sigma \big(\big[(\mathbf{I}-\mathbf{W}^T)^{-1} \mathbf{N}\big] \mathbf{A}^{(1)}\big) \mathbf{A}_j^{(2)}$, hence enabling the model to capture nonlinear relations between the observations $\mathbf{X}$ and the adjacency matrix $\mathbf{W}$. To learn $\mathbf{W}$, \cite{yu19daggnn} proposes an approach similar to that of variational auto-encoders (VAEs) \cite{kingma2014autoencoding}, where an encoder (another MLP followed by the linear transformation of $\mathbf{I}-\mathbf{W}^T$) encodes the input $\mathbf{X}$ into the latent variable $\mathbf{N}$, which is then passed through the decoder in Eq.~(\ref{eq:dag-gnn}). Unlike an implicit graph of functional dependence as in \cite{zheng20learning}, \cite{yu19daggnn} solves for an explicit DAG with adjacency matrix $\mathbf{W}$, by minimizing the evidence lower bound (ELBO) subject to the DAG constraint as in \cite{zheng18dags}. 

The encoder-decoder architecture in \cite{yu19daggnn} is a powerful framework for generative modeling, which we will discuss in more depth in Section \ref{sec:graph-generation}. Here we discuss another popular application of this framework in learning a data-driven graph topology from node observations. The main modeling assumption is that the temporal evolution of node observations, e.g., trajectories of interacting objects represented by $\mathbf{x}[t]$ for $t=1, \cdots, T$ and $\mathbf{X}$ in matrix form, is governed by a latent interaction graph. The goal is to recover this latent graph for explaining past and predicting future trajectories. A representative framework is \cite{kipf18neural}, in which an encoder takes $\mathbf{X}$ as input and returns a factorized distribution of $w_{ij}$ that represents the edge types between node pairs $i$ and $j$. A decoder then predicts $\mathbf{X}$ given the latent graph $\mathbf{W}$: 
\begin{equation}
\widehat{f}_\mathbf{X}: p_{\boldsymbol{\theta}}(\mathbf{X}|\mathbf{W}) = \prod_{t=1}^T p_{\boldsymbol{\theta}}(\mathbf{x}[t+1] \big| \mathbf{x}[t], \cdots, \mathbf{x}[1], \mathbf{W}).
\label{eq:nri}
\end{equation}
Under a Markovian assumption, $p_{\boldsymbol{\theta}}(\mathbf{x}[t+1] \big| \mathbf{x}[t], \cdots, \mathbf{x}[1], \mathbf{W}) = p_{\boldsymbol{\theta}}(\mathbf{x}[t+1] \big| \mathbf{x}[t], \mathbf{W})$ and \cite{kipf18neural} proposes to model it via a GNN. Taking a similar variational approach as in \cite{yu19daggnn}, the parameters associated with the encoder and decoder are trained by maximizing the ELBO. Building upon \cite{kipf18neural}, \cite{lowe22amortized} further tackles the challenges of amortized inference, i.e., learning potentially different DAGs for different time-series data at the same time, under the assumption of a shared underlying dynamical process. These examples demonstrate the significant flexibility offered by the encoder-decoder framework, which does not need to rely on a prototype dynamical or filtering process such as the SEM. 

In the examples discussed above, training is often facilitated via a loss function measuring how the learned graph helps reconstruct the observed data. Alternatively, in scenarios where ground truth graphs are available, they can be used together with the node observations to learn a mapping $\widehat{f}_\mathbf{X}$ directly. For example, \cite{ke23learning} considers a graph signal $\mathbf{x}$ as the outcome of an intervention performed on a certain node in a causal graph. Datasets collected from such interventions are combined with the ground truth causal graph to train a transformer-like architecture; once trained, the model can be used to reconstruct causal graphs from new interventional datasets. In a similar attempt, \cite{rossi22learning} considers node observations as equilibrium actions of a network game, whose utility function is, unlike in \cite{leng20learning}, not known a priori. This leads to a more general characterization of equilibrium actions satisfying: 
\begin{equation}
\widehat{f}_\mathbf{X}: \mathbf{x} = \mathcal{F}(\mathbf{W}) \mathcal{H}(\mathbf{b}), 
\label{eq:equi2}
\end{equation}
where $\mathbf{b}$ is the marginal benefit as in Eq.~(\ref{eq:equi}). Based on the relationship in Eq.~(\ref{eq:equi2}), \cite{rossi22learning} uses pairs of equilibrium actions $\mathbf{x}$ and networks $\mathbf{W}$ to train a transformer-like architecture, which implicitly captures the underlying utility function of the game. Once trained, the model can be used to reconstruct the network given new equilibrium actions. These two examples again demonstrate the flexibility of learning unknown network dynamics given the information at hand. 

\begin{figure}[t]
      \centering
        \includegraphics[width=16cm]{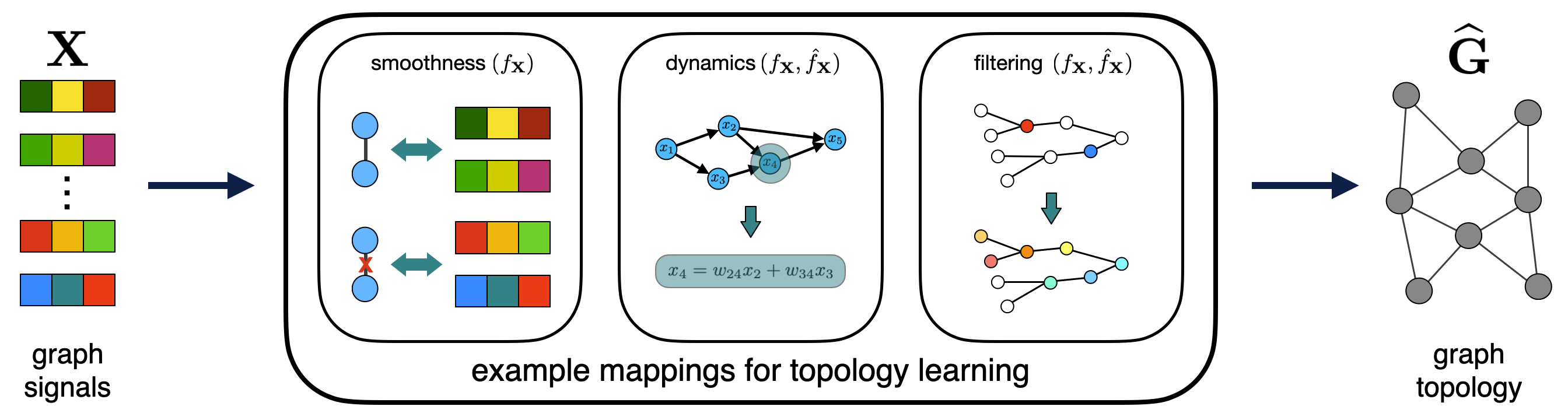}
        \caption{An illustration of methodologies for learning graph topology.}
        \label{fig:topology-learning} 
\end{figure}

\xd{A high-level illustration of the methodologies for learning graph topology can be found in Fig. \ref{fig:topology-learning}.} 

\subsection{Incorporating domain knowledge in topology learning}
\label{sec:topology-prior}

To conclude this section, we discuss the role of domain knowledge. As in many optimization or learning problems, prior or domain knowledge can provide useful inductive biases that guide the search toward desirable solutions, and the problem of graph topology learning is no exception. Various approaches have attempted to incorporate topological priors or constraints into the learning process. One of the most typical constraints is sparsity, which posits that the learned graph topology contains a relatively small number of significant edges. For example, this is promoted via the $\ell_1$-norm of the precision matrix in Eq.~(\ref{eq:glasso}) of the graphical lasso formulation, a combination of a logarithmic term on the node degrees and Frobenius norm of the adjacency matrix in \cite{Kalofolias16}, and explicitly specifying the number of desired edges in \cite{Chepuri17} and \cite{ying23network}. 
Another topological constraint that has attracted increasing interest is the  product-graph structure. For example, \cite{kadambari21product,shi24learning} consider a graph that can be factorized as the Cartesian product of two graphs, i.e., $G = G_P \oplus G_Q$, which implies $\mathbf{L} = \mathbf{L}_P \oplus \mathbf{L}_Q = (\mathbf{U}_P \mathbf{\Lambda}_P \mathbf{U}_P^T) \oplus (\mathbf{U}_Q \mathbf{\Lambda}_Q \mathbf{U}_Q^T)$, where $\oplus$ represents the Kronecker sum. From this, the same factor analysis model as in \cite{Dong16} is adopted, where $\mathbf{U}$ is replaced with $\mathbf{U}_P \otimes \mathbf{U}_Q$. This leads to the following distribution of $\mathbf{x}$:
\begin{equation}
f_\mathbf{X}: p(\mathbf{x}|\mathbf{L}) \sim \mathcal{N} \big( \mathbf{0}, (\mathbf{L}_P \oplus \mathbf{L}_Q)^\dagger + \sigma_N^2 \mathbf{I} \big). 
\label{eq:product}
\end{equation}
Based on this, \cite{kadambari21product} proposes to jointly solve for $\mathbf{L}_P$ and $\mathbf{L}_Q$ promoting smoothness on $G_P$ and $G_Q$, respectively, with additional sparsity and rank constraints. In a similar effort, \cite{lodhi20learning} further extends the setting above to learning the Kronecker product of two graphs. 
Extending beyond sparse and product graphs, \cite{kumar19structured} proposes to learn a $k$-component graph by enforcing the first $k$ eigenvalues of the learned graph Laplacian to be zero, \cite{rey23enhanced} enforces the learned graph to have a similar density of motifs compared to a given reference graph, and \cite{zheng18dags} learns a DAG by introducing a novel acyclicity constraint.  

The consideration of topological constraints above is motivated by both their prevalence in real-world networks, and the ease with which they can be explicitly encoded in closed forms amenable to optimization. 
With the mathematical ease, however, comes the restrictive set of constraints that may be considered. Important characterizations in network science, such as the distribution of node degrees or local clustering coefficients, are more difficult to encode and hence have been less studied. To address this, recent literature has started to investigate implicit ways to incorporate topological constraints. For example, in \cite{pu21learning}, the learned graph is encouraged to be similar to a template graph in a latent space constructed by a VAE, hence implicitly capturing topological properties of the template graph. Similarly, in \cite{sevilla24bayesian}, using a collection of ground truth graphs, a GNN is trained to output a score function that approximates the gradient of the log prior on the vectorized adjacency matrix, hence implicitly capturing any domain knowledge encoded by the ground truth adjacency matrices. Because these approaches directly make use of ground truth graphs, they can be viewed as data-driven ways of imposing topological constraints. The contrast between the explicit and implicit approaches hence mirrors that between the model-based and data-driven approaches discussed in Section \ref{sec:known-fX} and Section \ref{sec:unknown-fX}.

\section{Learning Graph Generation}
\label{sec:graph-generation}

\subsection{Learning graph generative models with a known mapping $h_{\bf G}$}
\label{sec:known-hG}
Having discussed the problem of learning graph topology from observed signals, we now turn to Problem \ref{prob:graph}, which focuses on learning how graphs themselves are generated. 
Specifically, we consider
a set of $M$ observed graphs $\{ G_i \}_{i=1}^M$, each endowed with the adjacency matrix $\mathbf{W}_i \in \mathbb{R}^{N \times N}$.  We assume a known graph generation mapping $h_{\bf G}: \Theta \to {\cal G}$ such that each $G_i$ can be treated as the realization of a distribution specified by $h_{\bf G}( \theta )$, where $\theta \in \Theta$ is the parameter for the distribution. The goal is to learn $\theta$ from the observed $\{ G_i \}_{i=1}^M$, which can then be used to generate new graph instances. 
From a statistical learning perspective, solving Problem \ref{prob:graph} is equivalent to solving a parameter estimation problem. 
In the following, we discuss examples of the known graph generation mapping $h_{\bf G}$ and the corresponding parameter estimation approaches, as well as their extensions. 

\begin{example}[\bf Erdős–Rényi (ER) model] \label{example:er}
This is one of the simplest models that can be used to describe `unstructured' graph topology. The parameter space is given by the set of positive real numbers $\Theta \equiv \mathbb{R}_{++}$ denoted by $\theta = p$. The $n$-node graph $G = (\mathcal{V},\mathcal{E}) \sim h_{\bf G}(p)$ has its node set fixed at $\mathcal{V} = \{1, \ldots, n\}$ and its edge set generated such that each edge is included in $\mathcal{E}$ with probability $\mathbb{P}( (i,j) \in \mathcal{E} ) = p$, independently of other edges \cite{kolaczyk2014statistical}.
\end{example}
The study of graph generative model learning for Example \ref{example:er} is instrumental. 
Since each edge is a Bernoulli random variable with the parameter $p$, the latter parameter can be straightforwardly estimated via the MLE principle, i.e., 
\begin{equation} \label{eq:er_mle}
    \widehat{p} = \frac{ \sum_{i=1}^m | \mathcal{E}(G_i) | }{ m \times {n \choose 2} },
\end{equation}
where $|\mathcal{E}(G_i)|$ denotes the cardinality of the edge set of the $i$th observed graph $G_i$. Note that the above is a consistent estimator such that $\widehat{p} \rightarrow p$ in mean-square sense as $m \to \infty$.

A closely related setup to the ER model is the $\beta$-model \cite{chatterjee2011random} motivated by random graphs with heterogeneous degrees. To represent the different degrees of nodes, the parameter space is extended to $\Theta \equiv \mathbb{R}_{++}^n$, denoted as $\theta = ( \beta_1, \ldots, \beta_n )$. The $n$-node graph $G = (\mathcal{V},\mathcal{E}) \sim h_{ \bf G }( \theta )$ is generated such that $\mathbb{P}( (i,j) \in \mathcal{E} ) = \frac{ e^{ \beta_i + \beta_j } }{ 1 + e^{\beta_i + \beta_j} }$, independently of other potential edges. As seen, $\beta_i$ increases the probability that node $i$ is connected to a neighboring node and thus the expected degree. Estimating the parameters via the MLE principle can be achieved by solving for $\{ \hat{\beta}_i \}_{i=1}^n$ in the following set of equations:
\begin{equation} \label{eq:mle_beta}
    \frac{1}{m} \sum_{s=1}^m d_i( G_s ) = \sum_{j \neq i} \frac{ e^{ \hat{\beta}_i + \hat{\beta}_j } }{ 1 + e^{ \hat{\beta}_i + \hat{\beta}_j  } },~i=1,\ldots,n,
\end{equation}
where $d_i(G_s)$ denotes the degree of node $i$ in the $s$th observed graph $G_s$. Note that \eqref{eq:mle_beta} can be solved using an efficient fixed-point iteration algorithm, see \cite{chatterjee2011random}, and the resultant estimator is consistent. A similar setup to the $\beta$-model can also be found in the Chung-Lu model \cite{chung2002connected} which follows a different parameterization for $\mathbb{P}( (i,j) \in \mathcal{E} )$.
Lastly, a generalized graph generative model that covers ER and $\beta$ models as special cases is the exponential random graph model (ERGM) \cite{lusher2013exponential}. The parameter space is given by $\Theta \equiv \mathbb{R}^d$. The $n$-node graph $G = (\mathcal{V},\mathcal{E}) \sim h_{ \bf G }( \theta )$ is generated such that
\begin{equation}
    \mathbb{P}( G ; \theta ) = \frac{ \exp( \theta^\top {\bf s}(G) ) }{ Z(\theta) },
\end{equation}
where ${\bf s}(G) \in \mathbb{R}^d$ is a vector of sufficient statistics that captures the topological properties of $G$, e.g., the number of edges, triangles, or other subgraph counts, and $Z(\theta) = \sum_{G \in {\cal G}} \exp( \theta^\top {\bf s}(G) )$ is the partition function. The MLE problem can be formulated as
\begin{equation}
    \max_{ \theta \in \Theta } ~\sum_{i=1}^m \log p( G_i ; \theta ) = \sum_{i=1}^m \theta^\top {\bf s}(G_i) - m \log Z(\theta) .
\end{equation} 
The challenge in solving the above MLE problem is that the partition function $Z(\theta)$ is intractable to compute, for which several approximation methods have been proposed \cite{chatterjee2013estimating, schweinberger2020concentration}. Moreover, \cite{schweinberger2020concentration} shows that the MLE problem is consistent under certain conditions on the sufficient statistics ${\bf s}(G)$.

Beyond learning the graph generation model directly from observed graphs, the literature has also considered an end-to-end learning problem provided that the mapping $f_{\bf X}$ is known and the only available data are $\{ \mathbf{x}_i \}_{i=1}^M$. The goal is to invert the composite function $f_{\bf X} \circ h_{\bf G}$ by sidestepping the inversion of $f_{\bf X}$, i.e., graph topology learning from graph signals in Section \ref{sec:known-fX}. Under the assumptions that $f_{\bf X}$ is related to a low-pass graph filter and that  $h_{\bf G}$ generates graphs with heterogeneous degrees, \cite{roddenberry2021blind} studied how to rank nodes according to the magnitude of their eigenvector centrality in an end-to-end learning fashion. Building on this setting, \cite{he2022detecting} proposed to apply factor analysis to relax the stationary graph signal assumption, while \cite{he2023online} extended the approach to inferring centrality from multiple graphs.

\begin{example}[\bf Preferential attachment (PA) model] \label{example:pa} 
    This is a model for describing graphs with power-law degree distributions \cite{barabasi1999emergence}. We consider the affine PA model in \cite{gao2017asymptotic}. The parameter space is given by $\Theta \equiv \mathbb{Z}_{++} \times (-1,\infty)$ with $\theta = (m, \delta) \in \Theta$. Here, $m$ is the number of edges to attach from a new node to existing nodes, and $\delta$ is the initial attractiveness of each node. The $n$-node graph $G = (\mathcal{V},\mathcal{E}) \sim h_{\bf G}(\theta)$ has its node set fixed at $\mathcal{V} = \{ v_1, \ldots, v_n\}$ and its edge set generated sequentially such that each new node $i$ connects to $m$ existing nodes with probability proportional to their degree plus $\delta$, independently of other edges. In detail, let ${\rm PA}_{t,i-1}(\delta)$ be the graph generated after adding the $(i-1)$-th edge for the $t$th node, and let ${\rm PA}_{0,0}(\delta) = {\cal G}_0$ be an initial graph. We include the $i$th edge to be added between the $t$-th node and $j$-th existing node ($j < t$) with the probability, 
    \begin{equation}
        \mathbb{P}( (v_t, v_j) \in \mathcal{E}( {\rm PA}_{t,i}(\delta) ) \, | \, {\rm PA}_{t,i-1}(\delta) ) = \frac{ d_j( {\rm PA}_{t,i-1}(\delta) ) + \delta }{ \sum_{k=1}^{t-1} ( d_k( {\rm PA}_{t,i-1}(\delta) ) + \delta ) },~\forall~j < t,
    \end{equation}
    where $d_j( {\cal G} )$ denotes the degree of node $j$ in graph ${\cal G}$.
\end{example}
For Example \ref{example:pa}, since exactly $m$ edges are added for every new node, the parameter $m$ can be directly computed from counting the number of edges in the observed graph. 
To estimate $\delta$, \cite{gao2017asymptotic} has derived and studied an estimator based on the MLE principle. In particular, \begin{equation}
    \widehat{\delta} = \arg\max_{ \delta > -1 }~ \left\{ \sum_{k=1}^\infty \log( k + \delta ) \frac{ N_{>k}(n) - N_{>k}^0(n) }{ n } - \frac{1}{2+\delta/m} \sum_{k=0}^\infty \log( k + 1 + \delta ) \frac{ N_{>k}^0(n) }{ n } \right\},
\end{equation}
where $N_{>k}(n)$ and $N_{>k}^0(n)$ are the numbers of nodes with degree larger than $k$ in the observed and initial graph, respectively. The estimator $\widehat{\delta}$ is shown to be consistent and asymptotically normal as $n \to \infty$.

\begin{example}[\bf Stochastic Block Model (SBM)] \label{example:sbm}
This is a popular model for describing graphs with block structure. The parameter space is given by $\Theta \equiv \mathbb{Z}_{++} \times \mathbb{R}_{+}^K \times \mathbb{R}_{+}^{K \times K}$ with $\theta = (K, {\bm p}, {\bm P} )\in \Theta$. Here, $K$ is the number of blocks, ${\bm p} \in \mathbb{R}_+^K$ is the block assignment probability vector with ${\bm p}^\top {\bf 1} = 1$, and ${\bm P} \in \mathbb{R}_{+}^{K \times K}$ is the matrix of edge probabilities between blocks. With this, the $n$-node graph $G = (\mathcal{V},\mathcal{E}) \sim h_{\bf G}(\theta)$ has its node set fixed at $\mathcal{V} = \{1, \ldots, n\}$. A block assignment vector ${\bm z} \in \{1,\ldots,K\}^n$ is then drawn according to  $\mathbb{P}( z_i = k ) = p_k$ for each $i \in \{1,\ldots,n\}$, and the edge set is generated such that each edge is included in $\mathcal{E}$ with probability $\mathbb{P}( (i,j) \in \mathcal{E} ) = P_{z_i , z_j}$, independently of other edges \cite{kolaczyk2014statistical}. Note that as a slight variation, the block assignment vector ${\bm z}$ can also be treated as a parameter to be learned, which is often the case in community detection problems.
\end{example}

For Example \ref{example:sbm}, we consider two sub-cases. First, when the block assignment vector ${\bm z}$ is known, the parameter ${\bm P}$ can also be estimated via MLE \cite{holland1983stochastic}, i.e., by estimating each entry $P_{k,l}$ by the proportion of edges between nodes in blocks $k$ and $l$ similar to \eqref{eq:er_mle}. Secondly, when the block assignment vector ${\bm z}$ is unknown (yet $K$ is known), \cite{snijders1997estimation} proposed an expectation-maximization (EM) method to approximate the corresponding MLE problem. Herein, the MLE problem is given by
\begin{equation} \label{eq:sbm_mle}
    \max_{ \theta \in \Theta } ~\sum_{i=1}^m \log p( \mathbf{W}_i ; \theta ) 
\end{equation}
with 
\begin{equation}
    p( \mathbf{W}_i ; \theta ) = \sum_{ {\bm z}_i \in \{1,\ldots,K\}^n } p( \mathbf{W}_i, {\bm z}_i; \theta ),
\end{equation}
\begin{equation}
    p( \mathbf{W}_i , {\bm z}_i; \theta ) = \prod_{j=1}^n p_{[{\bm z}_i]_j} \prod_{j<k} P_{ [{\bm z}_i]_j ,[{\bm z}_i]_k}^{ [{\bf W}_i]_{jk} } (1-P_{[{\bm z}_i]_j ,[{\bm z}_i]_k })^{1- [{\bf W}_i]_{jk} }.
\end{equation}
The summation in \eqref{eq:sbm_mle} is intractable as it involves $K^n$ terms. To remedy, \cite{snijders1997estimation} proposed to approximate the MLE problem by an EM method treating ${\bm z}$ as the latent variable. In particular, the E-step computes a set of sufficient statistics consisting of the conditional expectation of ${\bm p}$ and ${\bm P}$ given the current estimate of $\theta$ and the observed $\mathbf{W}_i$, and the M-step updates the estimates for $\theta$ using the sufficient statistics. Also see \cite{daudin2008mixture} for a similar variational algorithm and \cite{sussman2012consistent} which proved the consistency of such methods.
A popular alternative to EM or variational algorithms is to apply spectral clustering on each observed graph to identify the block assignment vector \cite{rohe2011spectral}, and then estimate the parameters similarly  to Eq.  \eqref{eq:er_mle}. In particular, \cite{su2019strong} showed that such an estimator is also consistent under certain conditions. 

An end-to-end learning approach has also been considered for SBM-like models. This leads to the \emph{blind community detection} problem, where the goal is to learn the block assignment vector ${\bm z}$ from observed graph signals $\{ \mathbf{x}_i \}_{i=1}^M$ without observing the graph topology $G$ (hence the term ``blind''). This problem can be treated as a special case of Problem \ref{prob:topo-task}, where the downstream task is to predict the block assignment vector ${\bm z}$ given $\mathbf{x}$. Surprisingly, the blind community detection problem can be handled by applying simple spectral clustering on the covariance matrix of graph signals \cite{wai2019blind, roddenberry2020exact}, with \cite{roddenberry2020exact} also providing theoretical guarantees for this problem under certain conditions.

\subsection{Learning graph generative models with an unknown $\widehat{h}_\mathbf{G}$}
\label{sec:unknown-hG}

We now consider the data-driven counterpart of Section \ref{sec:known-hG} in solving Problem \ref{prob:graph}.
Given a collection of observed graphs $\{G_i\}_{i=1}^{M}$, with adjacency matrices $\{\mathbf{W}_i\}_{i=1}^{M} \in \mathbb{R}^{N\times N}$, the goal is to learn an unknown graph generation mapping $\widehat{h}_\mathbf{G} : \Theta \rightarrow \mathcal{G}$, which represents a probability distribution over graphs. Once learned, $\widehat{h}_\mathbf{G}$ can be used to sample new graph instances, complete partially observed graphs, or learn latent representations useful for downstream prediction tasks. 

In contrast to model-based graph generative models such as the ER/PA models and SBM discussed in Section \ref{sec:known-hG}, where the form of $h_G$ is specified \emph{a priori}, data-driven graph generative models learn $\widehat{h}_\mathbf{G}$ directly from graph observations. This provides greater flexibility in modeling complex graph distributions, but typically comes at the cost of weaker identifiability and fewer statistical recovery guarantees.
Many data-driven graph generative models introduce a latent variable $z \in \mathcal Z$, from which a graph is generated according to
\begin{equation}
z \sim p(z), ~~~ G \sim p_\theta(G \mid z),
\end{equation}
where $p_\theta(G|z)$ is parameterized by a neural network. The latent representation may encode global graph-level structure, node-level embeddings, or both. Learning then amounts to estimating the parameters $\theta$ of the generative process from the observed graph collection $\{G_i\}_{i=1}^{M}$.

\begin{example}[\bf Graph variational auto-encoders (Graph VAEs)] \label{example:ivb-vae}

Graph VAEs learn a latent representation of each observed graph and decode it into a graph topology. Given an observed graph $G$, an encoder $q_\phi(z|G)$ maps $G$ to a latent variable $z$, sampled from a simple prior distribution, typically a standard Gaussian. 
A decoder $p_\theta(G|z)$ then  reconstructs or generates a graph from $z$. 

\end{example}

The model is typically trained by maximizing the ELBO, i.e., 
\begin{equation}
\mathcal L(\theta,\phi)
=
\mathbb E_{q_\phi(z|G)}
\left[
\log p_\theta(G|z)
\right]
-
D_{\mathrm{KL}}
\left(
q_\phi(z|G)\,\|\,p(z)
\right).
\end{equation}
The objective consists of two complementary terms. The first term, $
\mathbb E_{q_\phi(z|G)}
\left[
\log p_\theta(G|z)
\right],$
is the reconstruction term. It encourages the decoder to generate graph
instances that closely resemble the observed graph $G$ when sampled from
the latent representation $z$. Maximizing this term improves the fidelity
of the generated graphs. The second term, $
D_{\mathrm{KL}}
\left(
q_\phi(z|G)\,\|\,p(z)
\right),$
is the Kullback--Leibler (KL) divergence between the learned latent distribution and a chosen prior distribution, typically a standard
Gaussian \cite{kipf2016variational}. Minimizing this term encourages the latent representations inferred from observed graphs to remain close to the
prior distribution, thereby preventing the latent space from becoming overly specialized to the training data. As a result, nearby points in the latent space tend to generate similar graphs, yielding a smooth and structured latent representation that supports interpolation, sampling, and generation of previously unseen graph instances. Representative examples of this family include \cite{simonovsky2018}, \cite{pmlr-v80-jin18a}, \cite{vignac2022topn}. 

An important distinction from the representation-learning viewpoint discussed in Section III-A is the role of the latent space. In graph topology learning, latent variables are typically introduced to model the generation of graph signals conditioned on an underlying graph structure. By contrast, in graph VAEs, the
latent variable $z$ models the variability of the graph topology itself. The latent space therefore captures a distribution over graphs rather than a distribution over
signals observed on a graph. 
 
The latent-variable formulation adopted by graph VAEs provides a flexible mechanism for learning graph distributions. However, it does not explicitly model the sequential dependencies that may arise between graph components during graph generation. More generally, many classical graph generative models assume that edges are
generated independently, either globally as in ER models or conditionally on latent variables as in SBM. While such assumptions lead to tractable likelihood functions and efficient learning procedures, they can be restrictive when modeling real-world graphs, which
often exhibit higher-order dependencies such as clustering, motifs, and long-range dependencies. 

\begin{example}[\bf Autoregressive (AR) graph generative models] \label{example:ivb-AR}

Autoregressive graph generative models construct a graph through a sequence of conditional generation steps. Rather than generating the entire graph at
once, they factorize the graph distribution as
\begin{equation}
p_\theta(G)
=
\prod_{t=1}^{T}
p_\theta(a_t \mid a_{<t}),
\end{equation}
where $a_t$ denotes a generation action, such as adding a node, creating an edge, assigning an attribute, or terminating the generation process.
 
\end{example}

This formulation enables the model to capture complex dependencies between
graph components while naturally enforcing structural constraints during generation. Representative examples include GraphRNN \cite{conf/icml/YouYRHL18} and GraphAF \cite{shi2020graphaf}. While autoregressive models often produce high-quality samples when generating small to medium-size graphs, they remain sensitive to node ordering and their sequential
nature can lead to computationally expensive training and generation,
particularly for large graphs. More recently, hierarchical autoregressive formulations have been proposed to alleviate these limitations by generating graphs through multiscale expansion rather than individual node additions. For example, \cite{bergmeister2023efficient} introduces an algorithm based on iterative local expansion, which first generates a coarse graph representation and then progressively expands selected nodes into local subgraphs using a diffusion model. Instead of autoregressively predicting every node or edge, the method autoregressively refines graph resolution, combining hierarchical graph growth with local diffusion-based generation. This substantially improves scalability while preserving the expressiveness of diffusion models, as discussed next, enabling the generation of graphs with several thousand nodes and better extrapolation to unseen graph sizes. Along a different line of research, AutoGraph \cite{chen2026flatten} demonstrates that autoregressive graph generation can be formulated as a sequence modeling problem by introducing a reversible graph-to-sequence transformation. This enables scalable graph generation with standard decoder-only Transformers, effectively bridging graph generation and autoregressive language modeling.

\begin{example}[\bf Graph generative adversarial networks] \label{example:ivb-gans}
Generative adversarial networks (GANs) formulate graph generation as an adversarial learning problem between two neural networks: a generator that synthesizes graph instances from latent variables and a discriminator that aims to distinguish generated graphs from real ones. Unlike VAEs, GANs learn an implicit graph distribution without explicitly maximizing the data likelihood. Instead, the generator is trained to produce realistic graph structures that fool the discriminator, leading to a minimax optimization problem. 
\end{example}

More specifically, given a latent variable $z\sim p(z)$, the generator $G_\theta$ produces a graph
$\widehat{G}=G_\theta(z)$,
while the discriminator $D_\phi$ predicts whether a graph originates from the training data or from the generator. The two networks are trained by solving the adversarial objective
\begin{equation}
\min_{\theta}\max_{\phi}
\;
\mathbb{E}_{G\sim p_{\mathrm{data}}}
\left[\log D_\phi(G)\right]
+
\mathbb{E}_{z\sim p(z)}
\left[\log\left(1-D_\phi(G_\theta(z))\right)\right].
\end{equation}

Early graph GANs include \emph{MolGAN} \cite{de2018molgan}, which combines adversarial learning with reinforcement learning (RL) to generate small molecular graphs while optimizing desired chemical properties.  More recently, \emph{SPECTRE} \cite{pmlr-v162-martinkus22a} demonstrated that conditioning one-shot graph generators on spectral graph representations substantially improves their expressive power, highlighting the benefits of incorporating global structural information into implicit graph generative models.

Compared with autoregressive and variational approaches, GANs generate graphs in a single forward pass without requiring explicit likelihood estimation. However, adversarial training is often unstable, suffering from issues such as mode collapse and optimization difficulties, particularly in discrete graph domains. 


\begin{example}[\bf Diffusion-based graph generative models] \label{example:ivb-diffusion}

Graph diffusion models generate graphs through an iterative denoising process \cite{lai2025principles}.
Like graph VAEs, they introduce latent representations to model the variability
of graph structures. However, rather than learning a direct mapping from a latent
variable $z$ to a graph $G$, diffusion models define a sequence of latent graph
representations
$G_0, G_1, \ldots, G_T$,
where $G_0$ denotes an observed graph and $G_T$ corresponds to a highly corrupted
or noisy graph representation.  Thus, unlike classical generative models, which learn a direct mapping from noise to data, diffusion models generate samples through a sequence of refinement steps, gradually evolving coarse structures into fine-grained details.
\end{example}

The generative process is defined through a forward diffusion process
$q(G_t|G_{t-1})$,
which progressively corrupts the graph structure, and a learned reverse process
$p_\theta(G_{t-1}|G_t)$, which reconstructs the graph by iteratively removing noise. A key feature of diffusion models is that the forward diffusion process
$q(G_t|G_{t-1})$
is specified \emph{a priori} and is not learned from data. This process moves different graphs into the same
simple noise space, which is easy to sample from and serves as the starting point
for generation. Along the way, it also creates a smooth sequence of intermediate
noisy graphs connecting data to noise. 
Learning is therefore restricted to the reverse
process
$p_\theta(G_{t-1}|G_t)$,
which aims to reconstruct the graph by iteratively removing noise. Graph generation is
then performed by sampling from a simple prior distribution at time $T$ and
applying the learned reverse dynamics until a graph instance is obtained. We note that this reverse process does not recover one particular original graph. Instead, it starts from random noise and gradually turns it into a graph instance from the same underlying probability distribution.

Depending on the application, the diffusion process may act directly on the
adjacency matrix, on node and edge attributes, or on continuous latent graph
representations. 
A diverse family of graph diffusion models has consequently emerged. Early approaches include  score-based approaches~\cite{niu2020permutation, jo2022score, yan2023swingnn},  discrete diffusion~\cite{vignac2022digress, haefeli2022diffusion}, and geometric diffusion models for molecular conformation generation including DGSM \cite{NEURIPS2021_a45a1d12} and GeoDiff \cite{xu2022geodiff}.
They have also been employed as intermediate steps in specific generative schemes, such as hierarchical generation through iterative local expansion~\cite{bergmeister2023efficient}. More recent developments have extended diffusion to continuous-time formulations \cite{xu2024discretestate, siraudin2024cometh}, mixture diffusion models \cite{pmlr-v235-jo24b}, latent diffusion models \cite{yang2024graphusion, siraudin2026principled}, and structure-aware diffusion processes incorporating explicit graph priors such as SBMs \cite{pmlr-v267-su25e}. Diffusion has also been successfully combined with AR generation through approaches such as PARD \cite{zhao2024pard} and Arrow-Diff \cite{bernecker2024random}, illustrating the increasing convergence between sequential and diffusion-based graph generation paradigms. For an overview of graph diffusion models we point the reader to \cite{SHOU2026100854}.

In contrast to graph VAEs, where a single latent variable captures the graph distribution, diffusion models represent the generation process through a sequence of latent graph states. This iterative formulation often provides greater flexibility in modeling complex graph distributions and enables the
generation of high-quality graph samples. Compared with autoregressive approaches, diffusion models generate graphs
through iterative refinement rather than sequential graph construction,
allowing them to capture global graph structure more effectively.
Despite their empirical success, diffusion models typically require a large number of denoising steps during generation, resulting in substantial computational costs. This observation has motivated the development of flow-based and flow-matching approaches, which seek to directly learn the transport dynamics between simple and target graph distributions.

\begin{example}[\bf Normalizing flows and flow-matching graph generative models] \label{example:ivb-flowmatching}

Normalizing flows and flow-matching models constitute a family of likelihood-based generative approaches that learn a transformation between a simple reference distribution and the target graph distribution. Unlike autoregressive or diffusion models, which generate graphs sequentially or through iterative denoising, these methods explicitly model the probability transport from a tractable latent distribution to the graph space. Recent developments have further extended this paradigm from invertible transformations to continuous transport dynamics, establishing flow matching as an efficient alternative to diffusion-based graph generation.

\end{example}

Given an invertible transformation $f_\theta$, graph generation is performed by sampling a latent variable $z\sim p(z)$ and applying the inverse transformation $
G=f_\theta^{-1}(z)$,
while the exact likelihood is computed using the change-of-variables formula: 
\begin{equation}
\log p(G)=\log p(z)+
\log\left|
\det
\left(
\frac{\partial f_\theta(G)}{\partial G}
\right)
\right|.
\end{equation}

Early graph normalizing flows include \emph{GraphNVP} \cite{kaushalya2019graphnvp}, which introduced invertible coupling layers for molecular graph generation, and \emph{GraphAF} \cite{shi2020graphaf}, which combines autoregressive graph generation with flow-based transformations to improve scalability. Since graph generation often involves discrete node and edge attributes, subsequent work extended normalizing flows to discrete graph domains. \emph{GraphDF} \cite{pmlr-v139-luo21a} proposed discrete normalizing flows for molecular graph generation, while \cite{lippe2021categorical} generalized flow-based modeling to categorical variables through continuous transformations.

More recently, flow matching has emerged as a powerful alternative to classical normalizing flows. Instead of explicitly learning an invertible transformation and optimizing Jacobian determinants, flow matching directly learns the time-dependent transport dynamics between a simple reference distribution and the target graph distribution. Specifically, a neural vector field
\begin{equation}
\frac{dG_t}{dt}=v_\theta(G_t,t)    
\end{equation}
is learned to transport samples continuously from the source to the target distribution.
Following the general flow-matching framework of Lipman \emph{et al.} \cite{lipman2023flow}, and the generalization to discrete state space \cite{3692070.3692283}, \emph{DeFoG} was among the first methods to adapt flow matching to graph generation, demonstrating competitive sample quality with substantially fewer sampling steps than diffusion models \cite{pmlr-v267-qin25d}. 
In the meantime, variational flow-matching objectives have improved the stability and expressiveness of graph generation \cite{NEURIPS2024_15b78035}. 
More recent developments include constructing probability paths with smooth velocities that respect the graph geometry \cite{jiang26bureswasserstein}, 
interpreting conditional diffusion on graphs as a stochastic control problem with multiple reward signals \cite{tenorio26guided}, and principled transport maps for categorical variables extending flow-based generative modeling to discrete domains \cite{roos2026categorical}. 

Unlike diffusion models, which rely on a
fixed forward corruption process and a sequence of reverse denoising steps,
flow-matching models directly learn the continuous transport dynamics
between the reference and target distributions. This formulation can lead
to substantially faster sampling while preserving the ability to model
complex graph distributions.

While the previous Examples~\ref{example:ivb-vae}--\ref{example:ivb-flowmatching} focus on learning the empirical distribution of observed graphs, a different family of algorithms grounded in Bayesian generative approaches instead aims to learn sampling policies that approximate posterior distributions over graph structures conditioned on observed data. This alternative formulation is illustrated in Example~\ref{example:ivb-GenFlowNet}.

\begin{example}[\bf Bayesian graph structure generation via generative flow networks] \label{example:ivb-GenFlowNet}

Bayesian graph structure learning aims to characterize the posterior distribution over graph structures given observed data. Given observations $D$, the objective is to sample graph structures according to
\begin{equation}
p(G \mid D) \propto p(D \mid G)p(G),
\end{equation}
where $p(D|G)$ denotes the likelihood and $p(G)$ a prior over graph structures. Unlike the previous graph generative models, which aim to learn the empirical distribution of observed graphs, Bayesian graph structure learning seeks to characterize the posterior distribution over graph structures induced by the observed data. 

\end{example}

A recent line of work addresses this problem using generative flow networks (GFlowNets), with  representative examples including ~\cite{deleu2022bayesian}, ~\cite{deleu2023joint}. GFlowNets learn a stochastic policy that sequentially constructs a graph through a sequence of actions, such as adding an edge or assigning a parent node. Rather than maximizing the likelihood of observed graph instances, they optimize a flow consistency objective such that the probability of generating a graph is proportional to a prescribed reward. In Bayesian structure learning, this reward is naturally chosen as the unnormalized posterior, $R(G)=p(D|G)p(G)$, leading to a learned sampling policy satisfying $P_\theta(G)\propto R(G)$. Consequently, the learned policy directly approximates the posterior distribution over graph structures, enabling efficient sampling of multiple plausible graphs rather than producing a single point estimate.

Compared with classical Bayesian inference methods based on Markov chain Monte Carlo (MCMC), GFlowNets amortize the sampling process by learning a reusable policy, often resulting in substantially faster posterior exploration while maintaining sample diversity. Moreover, unlike graph VAEs, autoregressive models, diffusion models, and flow-matching approaches, which model the distribution of observed graph instances, GFlowNets are specifically designed to approximate posterior distributions over graph structures. This capability naturally supports Bayesian model averaging, uncertainty quantification, and robust structure discovery. 

\begin{figure}[t]
      \centering
        \includegraphics[width=16cm]{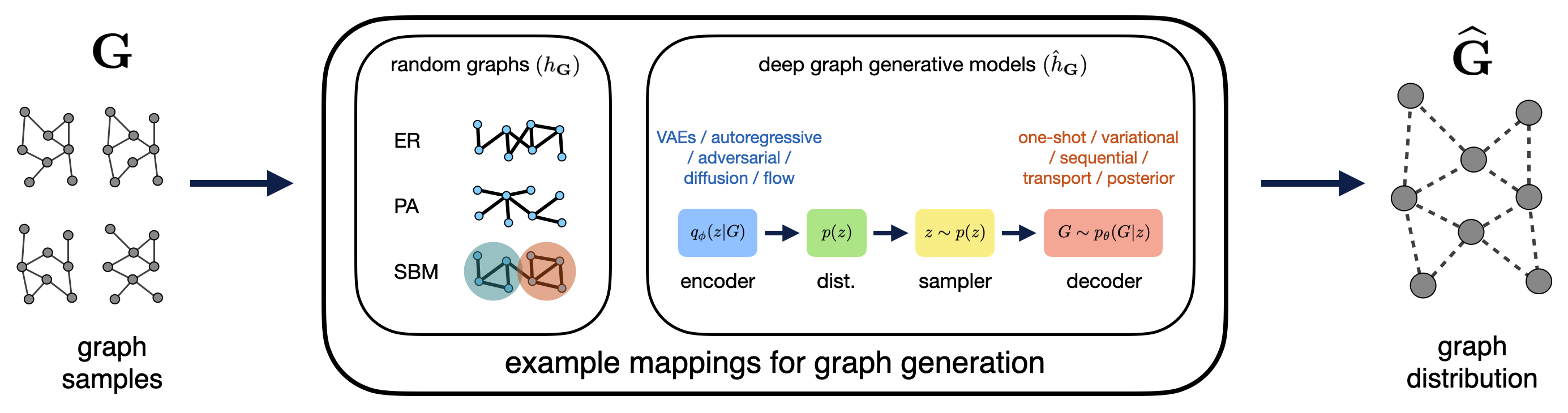}
        \caption{An illustration of methodologies for learning graph generation. The illustration of deep graph generative models is adapted from that in \cite{zhu22survey}.}
        \label{fig:graph-generation} 
\end{figure}

\xd{A high-level illustration of the methodologies for learning graph generation can be found in Fig. \ref{fig:graph-generation}.} 


\subsection{Incorporating domain knowledge in graph generation}
\label{sec:generation-prior}

Similar to topology learning, graph generation may also benefit from additional domain knowledge. While a known random graph model might have already specified the topological properties of the generated graph, 
a graph generative model trained from data may be constrained to learn or sample from a distribution whose support is restricted to admissible graphs. For example, PRODIGY introduces a plug-and-play projection mechanism that can be combined with pretrained continuous or discrete graph diffusion models. At each sampling step, the current graph representation is projected toward a user-specified constrained space, allowing hard and potentially non-differentiable requirements to be imposed without retraining the underlying generative model \cite{sharma2023plug}. ConStruct instead modifies the discrete diffusion process itself through an edge-absorbing noise model and a constraint-aware projector, ensuring that intermediate and final graphs remain within structural families satisfying properties such as planarity or acyclicity \cite{NEURIPS2024_f82385b8}. 
From this perspective, hard constraints play analogous roles in topology learning and graph generation: in the former they restrict the feasible graph estimate, while in the latter they restrict the support of the learned graph distribution. 

By contrast, there also exist attempts in which topological properties are absorbed from observed graph samples into the parameters or latent representation of a generative model, without defining an explicit feasibility set or topology-specific penalty. For example, NetGAN \cite{bojchevski2018netgan} learns the distribution of random walks on observed networks, thereby reproducing degree, community, and motif statistics without specifying these quantities explicitly. Similarly, GraphRNN \cite{conf/icml/YouYRHL18} and GRAN \cite{liao19efficient} generate graphs through conditional distributions over sequential graph-construction decisions, while SPECTRE \cite{pmlr-v162-martinkus22a} learns a distribution over spectral representations that encode global graph structure. Unlike projection-based approaches such as PRODIGY and ConStruct, these methods encourage topological validity only in distribution and do not guarantee that every generated sample satisfies the properties present in the training graphs. 

\section{Task-Driven Graph Structure Learning}
\label{sec:downstream-task}

The formulations presented in Sections \ref{sec:graph-topology} and \ref{sec:graph-generation} treat graph topology and graph generation as objectives in their own right. In many practical applications, however, the graph is not the final goal but rather an intermediate representation that supports a downstream task, such as node classification, link prediction, recommendation, or molecular property prediction. In these settings, the graph should be learned jointly with the downstream objective, leading to a task-driven formulation of graph structure learning, as defined in Problem \ref{prob:topo-task}.

\dt{Unlike the previous two formulations, task-driven graph structure learning does not simply invert either the $f$-map or $h$-map. Instead, the downstream objective introduces an additional supervision signal that jointly influences the learning process. From the perspective of our framework, the downstream task acts as an additional constraint on the inverse problem, biasing the learned graph topology or graph generative model toward representations that are useful for prediction rather than solely for explaining the observations. The additional supervision also changes the structure of the optimization problem: since the graph influences the downstream model, which in turn provides the supervision for learning the graph, the two components must be optimized jointly. In many cases, this mutual dependency naturally gives rise to a nested optimization problem, where the graph structure is updated according to the downstream objective, while the downstream model is simultaneously learned on the current graph.}

\begin{figure}[t]
      \centering
        \includegraphics[width=16cm]{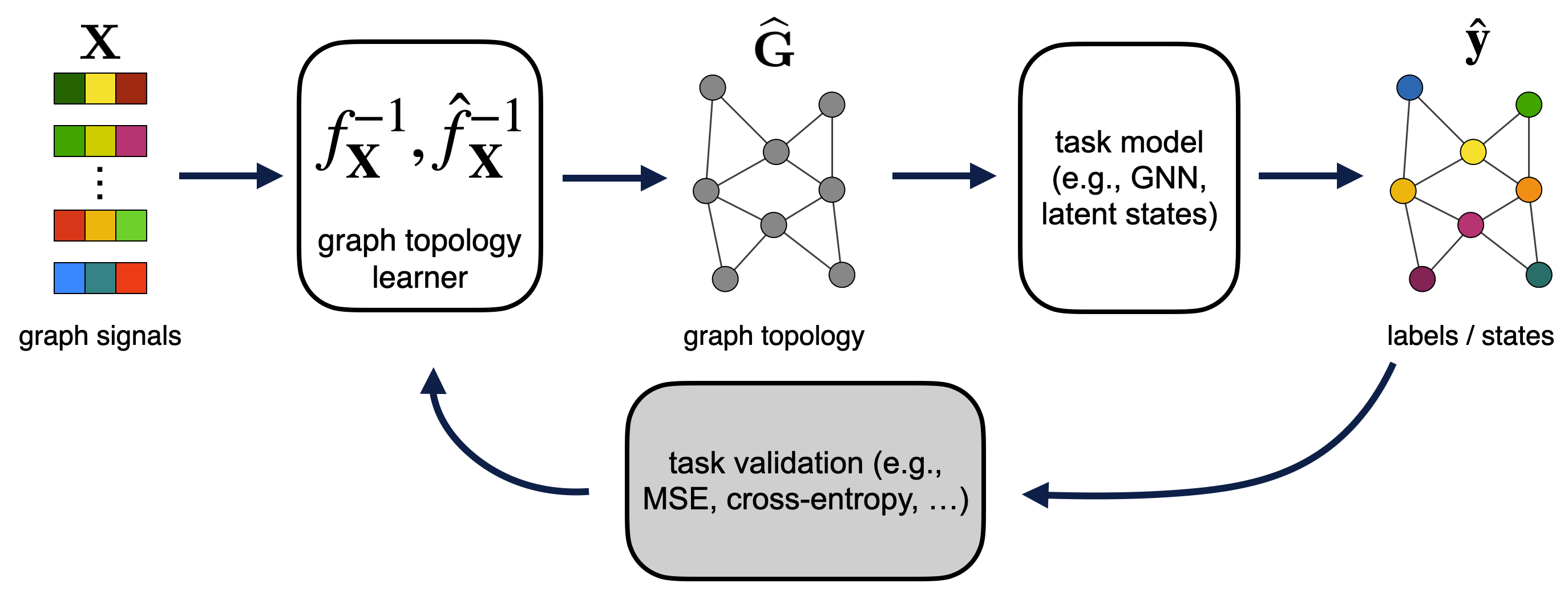}
        \caption{An illustration of task-driven graph topology learning.}
        \label{fig:task-driven} 
\end{figure}

\xd{Following the general idea outlined above, 
recent bilevel optimization (BLO) algorithms have played an important role in task-driven learning, particularly for the case of graph topology learning.} These advances are critical as the learning formulations involve implicit functions of the graph topology. In general, a BLO problem is described as:
\begin{equation} \label{eq:blo_general}
\begin{aligned}
\min_{ {\bf W}, {\bm\theta} } & ~~ J_{UL} ( {\bf W}, {\bm\theta} ) \\
\text{s.t.} & ~~ {\bm\theta} \in \arg\min_{ \hat{\bm\theta} } J_{LL} ( {\bf W}, \hat{\bm\theta} ),
\end{aligned}
\end{equation}
where $J_{UL} , J_{LL}$ are the upper-level (UL) and lower-level (LL) objective functions, respectively. 
A possible setup is that $J_{UL}(\cdot)$ models the performance of downstream tasks under the graph topology ${\bf W}$ and the auxiliary variable $\bm{\theta}$. The latter is learned by minimizing $J_{LL}(\cdot)$ that may incorporate the typical graph signal data.
Although \eqref{eq:blo_general} can be non-convex, recent works have developed efficient algorithms such as hypergradient-based methods \cite{ghadimi2018approximation} and penalty-based methods \cite{kwon2023fully}. These works have made it possible to handle the BLO that arises in task-driven graph topology learning.
In what follows, we review two examples of recent efforts in developing downstream task-induced topology learning techniques. 
A high-level illustration of the methodologies in this case can be found in Fig. \ref{fig:task-driven}.

\begin{example}[\bf Tasks through supervised graph neural networks]
In this example, the target downstream task is defined through the mapping $g_{\bf Y}: {\cal X} \times {\cal G} \to {\cal Y}$ where ${\cal Y}$ is a finite set describing the class labels on (a subset of) nodes.
As $y \in {\cal Y}$ represents the prediction target, the outputs of $g_{\bf Y}$ will be absorbed as a regularizer in the learning of the graph topology, i.e., affecting $J_{UL}(\cdot)$ in \eqref{eq:blo_general}. Meanwhile, the exact form of $g_{\bf Y}$ is typically unknown and is modeled as a GNN to be learned by minimizing a training loss in $J_{LL}(\cdot)$. 
\label{example:gnn}
\end{example}

To simultaneously learn $g_{\bf Y}$ and the graph topology that supports it, 
\cite{franceschi2019learning} proposed a bilevel optimization framework for learning graph structures that optimize the performance of GNNs on downstream tasks. The upper-level objective $J_{UL}(\cdot)$ measures the performance of the GNN on a validation set, while the lower-level objective $J_{LL}(\cdot)$ corresponds to the training loss of the GNN on a training set. Concretely,
\begin{equation} \label{eq:franceschi_form}
\begin{aligned}
& J_{UL} ( {\bf W}, {\bm\theta} ) = \mathcal{L} ( {\rm GNN}_{\bm{\theta}} ( {\bf W}, {\bf x}_{\rm val} ), {\bf y}_{\rm val} ), \\
& J_{LL} ( {\bf W}, {\bm\theta} ) = \mathcal{L} ( {\rm GNN}_{\bm{\theta}} ( {\bf W}, {\bf x}_{\rm train} ), {\bf y}_{\rm train} ),
\end{aligned}
\end{equation}
where ${\rm GNN}_{\bm{\theta}} ( {\bf W}, {\bf x} )$ denotes the output of a GNN defined on the graph ${\bf W}$ and input features ${\bf x}$ with parameters ${\bm\theta}$, and ${\cal L}(\cdot)$ is a standard training loss function (e.g., cross-entropy loss). The BLO problem is then tackled using a hypergradient-based method.
With a similar goal of leveraging downstream tasks to inform graph topology learning, \cite{jin2020graph} proposes Pro-GNN based on  a multi-objective formulation. This formulation can be interpreted  as a penalized BLO with $J_{LL}(\cdot)$ defined similarly to that in \eqref{eq:franceschi_form} and $J_{UL}(\cdot)$ chosen as in Section \ref{sec:known-fX}. The resulting BLO can be seen as a variant of \cite{franceschi2019learning}; also see 
\cite{zhu21survey,attali24rewiring,chen2022graph} for an overview.
The recent work in \cite{raghuvanshi2025task} further expanded the task-driven graph topology learning framework to incorporate learnable homophilic and heterophilic graphs, thereby increasing the  expressivity of the learned structures.
\htwai{Finally, we note that the GNNs are used in \eqref{eq:franceschi_form} as an example and they may be replaced by any graph machine learning model.}

\begin{example}[\bf Tasks through latent states]
In this example, the target downstream tasks are directly related to the functions performed by a graph system, e.g., graph signal denoising, sampling set selection, and utility maximization in an economic network. 
We can capture the downstream task by modeling the output of $g_{\bf Y}: {\cal G} \to {\cal Y}$ as a latent state. A typical setup for ${\cal Y}$ is the continuous space, e.g., ${\cal Y} = \mathbb{R}^{n \times d}$. The output of $g_{\bf Y}$ is incorporated as a regularizer in the learning of the graph topology in $J_{UL}$ of \eqref{eq:blo_general}. Meanwhile, the form of $g_{\bf Y}$ is typically assumed to be known and depends on the specific task.
\label{example:task}
\end{example}

Recent work \cite{zhang2026learning} studies graph topology learning in socioeconomic and biological systems by leveraging downstream task optimization. Specifically, \cite{zhang2026learning} models the downstream task through an equilibrium-seeking problem (e.g., a network game), which constitutes the LL problem in the BLO framework, and postulates that the graph topology results from a formation process that maximizes the utility evaluated at equilibrium.

The upper-level objective $J_{UL}(\cdot)$ measures simultaneously the fit of the learned graph topology to the observed graph signals and incorporates a regularization function depending on the equilibrium of the underlying network game or dynamical system  $\bm{\theta}^\star( {\bf W})$. The lower-level objective $J_{LL}(\cdot)$ characterizes its equilibrium condition. Concretely,
\begin{equation} \label{eq:zhang_form}
\begin{aligned}
\min_{ {\bf W}, \bm{\theta} } & ~~J_{UL} ( {\bf W}, {\bm\theta} ) = \mathcal{L} ( {\bf W}, {\bf X} ) + \lambda~\mathcal{R} ( {\bm\theta} ), \\
{\rm s.t.} & ~~\bm{\theta} \in \arg\min_{ \hat{\bm{\theta}} } J_{LL} ( {\bf W}, \hat{\bm\theta} ) = \mathcal{F} ( {\bf W}, \hat{\bm\theta} ),
\end{aligned}
\end{equation}
where $\mathcal{L}(\cdot)$ is a loss function measuring the fit of ${\bf W}$ to the observed graph signal data ${\bf X}$ [cf. Section~\ref{sec:graph-topology}], $\mathcal{R}$ is a regularization function that depends on the equilibrium of the underlying network game or dynamical system, and $\mathcal{F}$ characterizes its equilibrium condition. 
In a similar spirit, \cite{karaaslanli2022simultaneous} considers graph signal clustering as the downstream task and jointly learns the graph topology, \cite{batreddy2023robust} learns a graph representation to be used as a feature for classification, and \cite{sridhara2024towards} proposes to learn  the graph topology for sampling set selection.
By exploiting the structure of the task, these works also  show that a single-level formulation may suffice to approximately incorporate the downstream task as a direct regularizer in the objective of graph topology learning.

Unlike Example \ref{example:gnn}, where the downstream task is specified through a supervised GNN that can be directly learned, Example \ref{example:task} assumes that the downstream task is specified through a latent state arising from  \emph{known} dynamics. Works addressing unknown dynamics remain relatively scarce. The recent work in \cite{shao2026uniform} considers using observed trajectory data to learn the underlying dynamics that provide graph priors.
Research along this direction is still in its infancy, and further work is needed to fully understand the interplay between graph topology learning and downstream tasks. 
\xd{The discussion above also provides an interesting interpretation in which the task at hand effectively serves as the domain knowledge that may benefit learning. 
\cite{zhang2026learning} presents such an example, in which a viable social network topology is assumed to be one for which the associated equilibrium actions of a game played on that network would maximize the total utility of all individuals in the network (the task, which is reflected by the $\mathcal{R} ( {\bm\theta} )$ term in Eq.~(\ref{eq:zhang_form})). It is demonstrated that such an implicit bias can actually be interpreted using explicit topological properties, such as heterogeneity in node degrees, thereby bridging the gap between implicit and explicit graph priors.} 

Departing from graph topology learning, recent works have also considered learning graph generation models for downstream tasks. 
The underlying idea is similar, but the primary object to be learned becomes the graph generation model, e.g., a classical random graph model. For instance, for node classification tasks, \cite{zhang2019bayesian} proposes learning the posterior distribution of the parameters of an SBM from a potentially noisy graph and using it to sample new graph instances in an ensemble GNN model, while \cite{elinas2020variational} and \cite{kazi2022differentiable} propose learning a non-uniform ER model in which each edge is independently associated with a Bernoulli distribution. Graph diffusion models can also be adapted to downstream tasks. For example, \cite{xian2024graph} proposes a score-based diffusion model to perform graph classification rather than merely generating graph instances, \cite{tang2025cross} shows how to augment graph topology with a pre-trained diffusion model to serve various downstream tasks, 
and \cite{wang2025diffusion} considers augmenting a small graph with additional nodes generated by a latent diffusion model for node classification. 
Beyond the standard formulation of minimizing classification error, a related work \cite{you2018graph} considers an RL-based formulation to learn a policy network for graph generation, where the policy is directly influenced by a task-related reward such as synthetic accessibility or drug-likeness.

\section{Bridging topology learning with graph generation}
\label{sec:bridge}
Although topology learning and graph generation have been discussed separately, they share several fundamental characteristics. Recognizing these commonalities not only provides a higher-level synthesis of the two problems but may also inspire new formulations and methodologies. In this section, we summarize the commonalities and discuss some recent developments that bridge the two.

\subsection{A unified generating process for graph data}
\label{sec:V:unified}
Although topology learning and graph generation have traditionally evolved as distinct research directions, the framework illustrated in Fig. \ref{fig:paradigm} and presented in this review reveals that they are in fact complementary views of the same data-generating process. Indeed, topology learning focuses on recovering the latent graph structure from node observations by modeling the observation process $f_\mathbf{X}$. Under specific assumptions, such as smoothness, dynamical processes, or graph filtering, learning the graph simultaneously identifies the mechanism governing the observed signals. Consequently, the learned graph is often not only a structural object but also defines a generative model that explains the node observations, enabling, for example, the synthesis or prediction of graph signals according to the known or learned observation process. Conversely, graph generation assumes that graph instances are directly observed and aims to learn the graph-generating mechanism $h_\mathbf{G}$. Rather than estimating a single graph, these approaches learn a probability distribution over graphs, enabling the synthesis of previously unseen graphs with similar statistical and structural properties. Viewed together, these two families address complementary components of a single generative hierarchy, illustrated in Fig. \ref{fig:paradigm}, where latent generative parameters first determine graph structure, which subsequently gives rise to graph-structured observations through statistical, physical, or filtering processes. Existing methods typically recover only one stage of this hierarchy: topology learning focuses on the mapping between $\mathcal{G}$ and $\mathcal{X}$, while graph generation focuses on that between $\Theta$ and $\mathcal{G}$. Consequently, neither fully models the complete mechanism responsible for the observed graph data. 

Recent work has begun to bridge this gap by combining elements from the two traditionally separate stages discussed above. 
For example, \cite{wasserman23learning} formulates topology learning as a graph deconvolution problem, where the observed graph is modeled as the output of a latent graph passed through a graph convolutional operator. Therefore, this can be interpreted as learning a mapping $h_\mathbf{G}$ for $G$, which, however,  is inspired by a mapping $f_\mathbf{X}$ (i.e., convolutional filtering) commonly used to model node observations $\mathbf{x}$. Another example is the work in \cite{pmlr-v258-jiang25a}, which introduces an explicit probabilistic data-generating process for heterogeneous graphs and proposes to jointly learn graph topology together with the parameters governing graph formation, thus explicitly combining the mappings $h_\mathbf{G}$ and $f_\mathbf{X}$. 
A complementary line of work extends graph generation beyond the
topology alone. Graph-aware generative models such as
\cite{f03115b246304e0b8e453da5499c610e, uslu2026graph} incorporate graph operators into
the generation of graph-supported signals, thereby modeling not only
the structural object $G$ but also observations generated conditionally
on its geometry. These approaches illustrate two directions of
methodological transfer, leveraging the single generative hierarchy in Fig. \ref{fig:paradigm}: models developed for topology
learning can inform graph generation, while generative modeling can
extend topology learning from point estimates to probabilistic
models of structures and their associated observations.


\subsection{Bridging communities in graph structure learning }
\label{sec:V:communities}
The development of graph structure learning has been driven by several research communities that have historically evolved largely independently, including PGMs, GSP, GML, and network science. Although these communities often study similar graph-structured data, they have been motivated by different scientific questions, leading to the formulation of graph structure learning as an inverse problem. As a consequence, many methods that appear fundamentally different can instead be understood as learning different components of the same generating process for graph data introduced in Section \ref{sec:V:unified}, under different priors. 

At a high level, GSP views the graph as a latent physical or statistical structure that explains observed signals, i.e., a graph provides the domain over which the observations live. The primary objective is therefore to recover graph topologies that best support observed data under assumptions such as smoothness, diffusion, or dynamical evolution, as explicitly discussed in Section \ref{sec:known-fX}. In particular, most of the approaches rely on a notion of smoothness of the graph signal, either in the global sense (i.e., smoothness over all graph edges in Example \ref{example:iiia-smooth}) or in a more local sense (e.g., in Example \ref{example:iiia-dynamical} and diffusion-like graph filtering in Example \ref{example:iiia-filter}). The learned graph is expected to reflect meaningful interactions between variables and provide a data structure for downstream signal processing tasks including denoising, interpolation, forecasting, and system identification. As a matter of fact, a signal representation model is often employed and from this perspective graph topology learning can be understood as a form of representation learning for better processing of graph signals. For example, in the factor analysis model in Eq.~(\ref{eq:smooth}), the eigenvector basis can be viewed as a latent space (or an encoder) which is implicitly learned via the learning of $\mathbf{L}$, and the assumption on the signal $\mathbf{x}$ is mediated through that on its latent representation $\mathbf{h}$. Similarly, in the filter-based model in Eq.~(\ref{eq:filter}), the latent space becomes the eigenvector basis of $\mathbf{S}$ \cite{Segarra17a,Pasdeloup18}, or a dictionary that is a function of $\mathbf{S}$ \cite{Thanou17}. 

Similar to GSP, PGMs aim to recover interaction structures from observations, but define these interactions through conditional dependence relationships rather than graph signal models. Consequently, PGMs and GSP often arrive at closely related optimization problems despite originating from different modeling assumptions. There is also an analogy between classical PGM and dynamical-process-based approaches, where the graph captures the conditional dependence (bi-directional in PGMs) or conditional probability (directional in diffusion) between a node pair. Both can then be interpreted using a unified GSP language.

By contrast, GML primarily views the graph as a latent variable for representation learning. Rather than explicitly recovering the underlying interaction network, the objective is to learn representations that optimize generative or downstream predictive tasks. While GML has traditionally focused on discriminative problems such as node classification, graph classification, and link prediction, recent advances have increasingly focused on the specific problem of graph generation, where the objective is to learn expressive distributions over graph-structured data for applications including molecule design, simulation, and data augmentation. Most of the state-of-the-art methods in this category follow an encoder-decoder paradigm, sharing a similar philosophy to GSP approaches for topology learning. However, the latent space (encoder) can often be learned in a data-driven manner (note that \cite{pu21learning} already leverages this idea to implicitly incorporate graph priors), or predefined based on a noise model (see Example \ref{example:ivb-diffusion}).

A third perspective arises from network science and statistical network modeling, where the primary objective is to understand the principles governing network organization. Instead of learning a single graph instance, these approaches seek probabilistic mechanisms capable of explaining graph populations through properties such as community organization, degree distributions, motif statistics, or temporal evolution. Classical random graph models, SBMs, and PA models exemplify this viewpoint by treating graphs as realizations of underlying probabilistic mechanisms (i.e., $p(G|\Theta)$)  that explain why observed networks exhibit particular structural characteristics. These generative models are usually selected according to the context of the target complex systems under study. For example, the ER model (Example \ref{example:er}) is used to model random networks, the SBM (Example \ref{example:sbm}) is used to model community structures in social networks, and the PA model (Example \ref{example:pa}) is used to model scale-free networks in biological systems. Going beyond standard random graph models, recent studies have considered task-induced graph models \cite{zhang2026learning}, which are used to motivate the prior knowledge of the graph topology in the context of graph structure learning. 

Viewed through the lens of the unified generating process for graph data, these communities differ primarily in the level of abstraction at which learning is performed. GSP and PGMs seek interaction structures that explain observed data, GML learns representations and graph structures that optimize downstream predictive or generative objectives, while network science seeks the latent mechanisms governing graph formation itself. Thus, these perspectives address complementary learning problems defined on different components of the same generating process. See Table \ref{tab:communities} for an overview. Recognizing these connections provides opportunities to transfer ideas across communities by, for example, combining physically grounded graph priors from GSP with the flexibility of GML, or enriching deep graph generative models with the mechanistic principles developed in network science. \xd{Similarly, the divide between model-based and data-driven approaches is often less clear than that idealized in Fig. \ref{fig:paradigm}, and recent developments often leverage the strengths of both for the task at hand, with typical examples including learned priors \cite{pu21learning,sevilla24bayesian}, or constraint-aware generators \cite{sharma2023plug,NEURIPS2024_f82385b8}.}

\begin{table*}[t]
\centering
\caption{A unified view of major communities in graph structure learning.}
\label{tab:communities}
\renewcommand{\arraystretch}{1.25}
\scalebox{0.88}{
\begin{tabular}{|p{2.5cm}|p{2.5cm}|p{2.5cm}|p{4.5cm}|p{4.5cm}|}
\hline
\textbf{Perspective} &
\textbf{Observations} &
\textbf{Object of learning} &
\textbf{Representative tasks} &
\textbf{Representative models / methods} \\
\hline

Probabilistic Graphical Models (\ref{sec:known-fX}) &
Data samples $\mathbf{X}$ &
Conditional dependence graph $G$ &
Structure learning, covariance estimation, causal discovery &
Gaussian graphical models, Bayesian networks \\
\hline

Graph Signal Processing (\ref{sec:known-fX}, \ref{sec:unknown-fX}, \ref{sec:downstream-task}) &
Graph signals $\mathbf{X}$ &
Signal domain $G$ &
Topology learning, denoising, interpolation, forecasting &
Smoothness priors, graph filtering, dictionary learning \\
\hline

Network Science (\ref{sec:known-hG}) &
Graph samples $G$ &
Graph-generating mechanisms $p(G)$ &
Community detection, motif discovery, network evolution &
ER, ERGM, PA, SBM \\
\hline

Graph Generation (\ref{sec:unknown-hG}, \ref{sec:downstream-task}) &
Graph samples $G$ &
Generative model $p(G)$ &
Graph synthesis, simulation, uncertainty quantification &
VAEs, GANs, normalizing flows, diffusion models, ARs \\
\hline

Graph Machine Learning (\ref{sec:unknown-fX}, \ref{sec:unknown-hG}, \ref{sec:downstream-task}) &
Graphs $G$, node attributes $\mathbf{X}$ &
Predictive function $p(y|G,\mathbf{X})$ &
Node/graph classification, regression, link prediction &
Graph neural networks, graph transformers \\
\hline

\end{tabular}}
\end{table*}

\section{Future perspectives}
\label{sec:perspective}

The unified perspective presented throughout this review naturally gives rise to a new generation of formulations of graph structure learning that jointly model multiple components of the generating process for graph data. Rather than treating graph topology learning and graph generation as separate problems, these approaches seek unified models capable of explaining how relational systems emerge, evolve, and generate observations.

\subsection{From graphs with pairwise relations to higher-order structures and richer geometries}

Throughout this review, we have focused on learning the structure of pairwise graphs. However, many real-world systems exhibit interactions that cannot be adequately captured by pairwise relationships alone, suggesting that graphs may not always constitute the most appropriate relational representation. This motivates extending graph structure learning beyond pairwise graphs to higher-order structures such as hypergraphs \cite{tang2023learning,tang25markov,brown25scalable,duta2025sphinx}, simplicial complexes \cite{buciulea24learninggraphs,buciulea24learningtopology,gurugubelli24simplicial}, and cell complexes \cite{battiloro24latent}, as well as richer geometries including cellular sheaves \cite{bodnar23neural,dinino24learning,dinino26learning}, vector bundles \cite{bamberger25bundle}, and product manifolds \cite{borde23neural}. 

More fundamentally, future systems for graph structure learning may treat the choice of relational representation itself as part of the learning problem. Rather than assuming a priori that observations arise from a graph, learning algorithms could infer whether pairwise graphs, hypergraphs, simplicial and cell complexes, or richer geometries provide the most appropriate representation of the underlying interactions and their associated feature transformations. From this perspective, 
graph topology learning naturally generalizes from recovering an unknown graph to discovering a general topological or geometric domain, while graph generation extends from learning generative distributions over graphs to learning generative distributions over these richer representations. More broadly, this evolution suggests that the unified perspective developed throughout this review may itself be viewed as a special case of a broader paradigm: learning relational representations directly from data, where the underlying domain geometry is no longer prescribed but inferred.


\subsection{From learning a graph structure to learning a generating process}
\label{secVII:B}

The unified perspective developed throughout this review suggests a broader objective for graph structure learning. Rather than treating topology learning, graph generation, and downstream tasks as separate problems, they could be viewed as complementary components in a common relational data-generating process as discussed in Section \ref{sec:V:unified}. The mechanisms identified through such a holistic approach would shed light on how relational structures are formed, evolve, and generate observations. Under this perspective, the graph becomes an intermediate latent representation rather than the final object of learning.

Recent developments have begun to move in this direction. Probabilistic formulations of graph topology learning model the observed graph through latent graph-generating processes, 
thereby explicitly reasoning about how graph structures arise \cite{pmlr-v258-jiang25a}. Likewise, emerging approaches in causal structure learning employ structured search and RL to infer the mechanisms governing directed dependencies rather than merely estimating connectivity patterns \cite{aaai.v37i9.26274}. Although these first methods remain specialized, they illustrate a broader shift from estimating relational structures toward explaining the processes that generate the observations.

\subsection{Toward a universal learning paradigm via foundation models}

As discussed so far, existing graph topology learning and graph generation methods are typically designed for specific input data and modeling assumptions, graph domains, or downstream tasks. As graph-structured data continue to grow in scale and diversity, an important future direction is to develop generalist models capable of learning transferable representations and structural priors across heterogeneous graph learning problems. Recent advances in foundation models open new opportunities for integrating probabilistic learning with semantic and domain knowledge. For example, large language models have recently been leveraged to guide causal graph discovery by reducing the combinatorial search space and incorporating prior knowledge into the learning process \cite{jiralerspong2024efficient}. More broadly, foundation models may provide a common interface for combining graph signal processing, graph generative models, causal inference, and domain knowledge within a unified probabilistic learning framework, enabling graph learning systems that are more data-efficient and transferable across diverse application domains.

\subsection{Promising applications}

\xd{Although we have intentionally focused on the technical aspects of graph topology learning and graph generation in this review, they naturally enable numerous applications in biological, chemical, social, and urban networks, to name a few. Looking five to ten years ahead, we believe the most consequential applications of these tools are likely to emerge in systems where relational structure is latent, time-varying, and directly designable. On the one hand, graph topology learning may evolve from the offline recovery of a single static graph toward uncertainty-aware, multiscale inference of changing interactions from streaming observations, providing a structural foundation for self-updating physical infrastructure, industrial processes, autonomous collectives, social networks, and biomedical mechanisms. Existing work on graph-based physical simulation already illustrates the potential to learn interaction structure jointly with system dynamics \cite{sanchezgonzalez2020simulate}, while large-scale graph-based forecasting demonstrates the broader opportunity for relational models in complex dynamical systems \cite{lam2023graphcast}. On the other hand, deep graph generation may move beyond reproducing observed graph statistics toward goal-oriented structural design, generating proteins, molecular complexes, materials, and network configurations with specified functional and physical properties \cite{watson2023rfdiffusion,zeni2025mattergen}. Together, the most transformative systems may combine the two paradigms in a closed loop: inferring the current relational structure from observations, generating promising candidate structures for intervention, evaluating them through simulations or experiments, and updating both the inferred topology and the generative model from feedback.}

\bibliographystyle{IEEEtran}
\bibliography{mybibfile.bib}

\end{document}